\pdfoutput=1
\documentclass[11pt]{article}

\usepackage[]{ACL2023}
\usepackage{times}
\usepackage{latexsym}
\usepackage[T1]{fontenc}
\usepackage[utf8]{inputenc}
\usepackage{microtype}
\usepackage{inconsolata}
\usepackage{xcolor}
\usepackage{amsmath}
\usepackage{bbm}
\usepackage{graphicx}
\usepackage{booktabs}
\usepackage{float}
\usepackage{subcaption}
\usepackage{tikz}

\title{Dramatic Conversation Disentanglement}

\author{Kent K. Chang, Danica Chen \and David Bamman \\
       University of California, Berkeley\\
       \texttt{\{kentkchang,danicachen2019,dbamman\}@berkeley.edu}}

\definecolor{navyblue}{rgb}{0.0, 0.0, 0.5}
\definecolor{napiergreen}{rgb}{0.16, 0.5, 0.0}
\definecolor{neonfuchsia}{rgb}{1.0, 0.25, 0.39}
\definecolor{orange-red}{rgb}{1.0, 0.27, 0.0}
\definecolor{dimgray}{rgb}{0.41, 0.41, 0.41}
\definecolor{chromeyellow}{rgb}{1.0, 0.65, 0.0}
\usetikzlibrary{arrows.meta,backgrounds,positioning,fit,calc,intersections,shapes,patterns}

\newcommand{\vect}[1]{\boldsymbol{#1}}

\begin{document}
\maketitle
\begin{abstract}
We present a new dataset for studying conversation disentanglement in movies and TV series. 
While previous work has focused on conversation disentanglement in IRC chatroom dialogues, movies and TV shows provide a space for studying complex pragmatic patterns of floor and topic change in face-to-face multi-party interactions. 
In this work, we draw on theoretical research in sociolinguistics, sociology, and film studies to operationalize a conversational thread (including the notion of a floor change) in dramatic texts, and use that definition to annotate a dataset of 10,033 dialogue turns (comprising 2,209 threads) from 831 movies.  
We compare the performance of several disentanglement models on this dramatic dataset, and apply the best-performing model to disentangle 808 movies. 
We see that, contrary to expectation, average thread lengths do not decrease significantly over the past 40 years, and characters portrayed by actors who are women, while underrepresented, initiate more new conversational threads relative to their speaking time.
\end{abstract}

\section{Introduction}

Movie and TV dialogues, or dramatic dialogues more generally, have offered linguists a wealth of resources to study conversational behaviors~\cite{Lakoff1984-og,He1998-dk,Richardson2010-zp}, including within NLP~\cite{danescu-niculescu-mizil-lee-2011-chameleons,ramakrishna-etal-2017-linguistic,sap-etal-2017-connotation,azab-etal-2019-representing}. 
While dramatic dialogues do not necessarily mimic conversations in real life, they present complex pragmatic and sociolinguistic phenomena that warrant study; given the widespread viewership of movies and TV, what appears on screen---both visually and in dialogue---can have a real social impact in the world~\cite{rosen73,hooks92,Heldman2016}.

\begin{figure}[htbp!]
  \centering
    \begin{tikzpicture}[y=-0.5cm, %
                        line/.style={stealth-, out=270,in=130,thick},%
                        every node/.style={
                           outer sep=1pt,
                           text width=7.4cm,
                           font=\scriptsize\sffamily
                        }]
        \node[anchor=north west, color=navyblue] (a) at (0,0) {\textsc{\textbf{georgie}.} Morning.};
        \node[below=1mm of a.west, anchor=north west, color=navyblue] (b) {\textsc{\textbf{georgie sr.}} How's the ankle?};
        \node[below=1mm of b.west, anchor=north west, color=navyblue] (c) {\textsc{\textbf{georgie}.} I will be all right. Think I will be able to start \\ against Nacodoches?};
        \node[below=2.5mm of c.west, anchor=north west, color=navyblue] (d) {\textsc{\textbf{georgie sr}.} I can't play favorites Georgie, depends on how \\ hard you work.};
        \node[below=2.5mm of d.west, anchor=north west, color=napiergreen] (f) {\textsc{\textbf{missy}.} Mom, Sheldon can't find his bowtie.};
        \node[below=1mm of f.west, anchor=north west, color=napiergreen] (g) {\textsc{\textbf{mary}.} Really? I laid it out for him.};
        \node[below=1mm of g.west, anchor=north west, color=napiergreen] (i) {\textsc{\textbf{georgie sr}.} Leave it alone Mary, he doesn't need a damn bowtie.};
        \node[below=2.5mm of i.west, anchor=north west, color=napiergreen] (j)  {\textsc{\textbf{mary}.} It's his first day of school, let him wear what he wants.};
        \node[below=1mm of j.west, anchor=north west, color=neonfuchsia] (k) {\textsc{\textbf{sheldon} (o.s.).} MOM, I CAN'T FIND MY BOWTIE!!!};
        \node[below=1mm of k.west, anchor=north west, color=neonfuchsia] (l) {\textsc{\textbf{mary}} Oh dear Lord, why's he gotta wear a bowtie?};
        \node[below=1mm of l.west, anchor=north west, color=chromeyellow] (n) {\textsc{\textbf{georgie}.} Can I drive in with you?};
        \node[below=1mm of n.west, anchor=north west, color=chromeyellow] (o) {\textsc{\textbf{georgie sr}.} Sure.};
        \node[below=1mm of o.west, anchor=north west, color=chromeyellow] (p) {\textsc{\textbf{mary}.} Everybody's gonna know he's your brother.\\ You can't hide. It's gonna be awful for you.};        
        \draw [line,color=navyblue] (a.west) edge (b.west);
        \draw [line,color=navyblue] (b.west) edge (c.west);
        \draw [line,color=navyblue] (c.west) edge (d.west);
        \draw [line,color=napiergreen] (f.west) edge (g.west);
        \draw [line,color=napiergreen] (g.west) edge (i.west);
        \draw [line,color=napiergreen] (i.west) edge (j.west);
        \draw [line,color=neonfuchsia] (k.west) edge (l.west);
        \draw [line,color=chromeyellow] (n.west) edge (o.west);
        \draw [out=220,in=150,thick,color=chromeyellow] (n.west) edge (p.west);
    \end{tikzpicture}
  \caption{
  Example of dramatic conversations, taken from a scene in \textit{Young Sheldon}. 
  Speaker labels are in boldface and~\textsc{small caps}.
  Curved arrow lines indicate the reply-to relations between dialogue lines.
  Each thread is distinguished by colors.
  }
  \label{fig:intro-example}
\end{figure}
An important feature of such dialogues is that they are \textit{entangled}. 
In his work on dramatic dialogues, \citet[p. 3]{McKee2016-fo} articulates a speech act view: in screenplays, ``all talk responds to a need, engages a purpose, and performs an action.''
In any given scene in a movie or TV show, then, we can often see multiple needs expressed by different characters in one sequence of conversation.
Consider this scene from \textit{Young Sheldon}~(Fig.~\ref{fig:intro-example}):
Missy relays a message, Sheldon wants to know where his bow tie is, and Georgie seeks to avoid showing up with Sheldon at school---each of them starts a new conversational thread (or, subconversation) with their speech act that reflects those different intents.
In the screenplay, there is no explicit structure that indicates where each subconversation starts and ends.
If, however, we could disentangle dramatic conversations, we could ask such questions as: 
What kind of characters get to \emph{start} a thread? 
How long do conversations tend to last? 
Answers to those questions can enhance our understanding of cultural representations on screen. %

Much of the work on conversation disentanglement in NLP has studied Internet Relay Chat (IRC) logs, most notably~\texttt{\#Linux}~\cite{elsner-charniak-2008-talking} and~\texttt{\#Ubuntu}~\cite{kummerfeld-etal-2019-large}. %
IRC logs are in a different domain from dramatic conversations, and some salient features in IRC, such as invoking a username to indicate replies to that user, are not found in screenplays.
Conversely, there is no equivalent for ``off-screen'' speakers in IRC.
Given the face-to-face nature of conversations in drama, movie and TV characters can start new conversational threads by entering the scene. %
Those major differences mean chat logs may be insufficient to train models that disentangle drama.

To bridge the gap, we present in this work a new annotated dataset to support the study of conversation disentanglement in the domain of movies and TV shows. 
We draw heavily on the theoretical resources found in film studies, sociology, and linguistics as we design our annotation framework, with particular attention paid to the semantic and pragmatic signals of the start of a new thread.

In this work we make the following contributions:%

\begin{itemize}
    \item We draw on theoretical research in sociolinguistics, sociology, and film studies to operationalize a conversational thread (including the notion of a floor change) in dramatic texts, and use that definition to annotate a dataset of 10,033 dialogue turns (comprising 2,209 threads) from 831 movies. All annotations are freely available for public use under a CC BY-NC-SA license on GitHub.\footnote{\url{https://github.com/kentchang/dramatic-conversation-disentanglement}} 
    \item We compare the performance of several disentanglement models on this dramatic dataset to see if model architectures designed for or models trained on~\citet{kummerfeld-etal-2019-large} perform well in the domain of drama. 
    \item We apply the best-performing model to analyze and disentangle 808 films in \textsc{Scriptbase-J}~\cite{gorinski-lapata-2015-movie,Gorinski2018-rh}, investigating both the relationship between historical thread length and intensified continuity style~\cite{Bordwell2002-rn} and the relationship between gender and power in floor claiming. In this data, we see that, unlike shot lengths, average thread lengths do not decrease significantly over the past 40 years (contrary to expectation), and characters portrayed by  actors who are women, while underrepresented, initiate more new conversational threads relative to their speaking time.
\end{itemize}

\section{Related work}

\paragraph{Conversation disentanglement.} Conversation disentanglement seeks to identify threads (or, clusters, subconversations) in a sequence of utterances.
Conceptually, this task requires a robust operationalization of \textit{thread}, which is usually understood as related to topic or floor change~\cite{ONeill2003-bf,Shen2006-qc,Jiang2018-gp}. 
\citet{elsner-charniak-2008-talking,Elsner2010-wi} considered this problem in the context of chat history, which has been extensively studied since~\citep[for a recent survey, see][]{Gu2022-hj}. 
 
In terms of modeling, there are two popular approaches for this task~\cite{Zhu2021-yq}: two steps (models link individual utterances first, and then we recover thread membership) or end-to-end (models predict thread membership directly). 
Our work adopts the two-step method: we first calculate the similarity score to identify the reply-to relations between two utterances (the link prediction task), and apply a greedy clustering algorithm to put utterances that reply to one another into threads (the clustering task). 
For the two-step method, there have been attempts to adopt a multi-task learning setup:
At training time, when gold cluster information is available, this auxiliary task calculates another loss function dedicated to thread prediction, which can be used to improve the performance of link prediction, the main task, with which we also experimented. 

Datasets for conversation disentanglement are currently limited. 
In \citeposs{Mahajan2021-ov} comprehensive survey on multi-party dialogue understanding, most datasets are not curated with this purpose in mind. 
\citet{kummerfeld-etal-2019-large}'s corpus, built on annotations of IRC chat logs, has been the standard benchmark dataset for this task. 
\citet[p. 3871]{Liu2020-rr} released a dataset of movie dialogues, where they ``collect 869 movie scripts that explicitly indicate the plot changing'' and ``extract 56,562 sessions from the scripts and manually intermingle these sessions to construct a synthetic dataset.''  
In this work, we present a new annotated dataset built on movies and TV shows, adding a spoken, scripted corpus to facilitate this line of work. 
Instead of inferring from the narrative structure and threading conversations through a synthetic process, we developed an annotation framework (described in \S\ref{sec:anno}).

\paragraph{Theoretical approaches to conversation.} Conversation has been extensively theorized and studied in sociology, linguistics, and film studies, which this work draws on.
For our annotation scheme, \citeposs{Goffman1963-ze} idea that conversation is a form of focused interaction and~\citeposs{McKee2016-fo} speech act view on dramatic dialogues provide us with the theoretical foundation.

The ideas related to \textit{focus} and \textit{topic} are further explored in the following: 
\citet{Ervin-Tripp1964-dq}, who considers the surface and semantic features of sustained attention in conversational organization;~\citet{Roberts1996-tc}, who expounds on ``topic under discussion'' in pragmatics;~\citet{Ng1993-ju}, who see topic change and floor claiming from the perspective of power dynamics between speakers involved.
More broadly,~\citet{Sacks1974-cg,Goodwin1981-wq} detail the organizational and pragmatic principles for conversation, and~\citet{He1998-dk} consider them specifically in the context of drama.

This work is further motivated by the social implications of conversations taking place on screen, and we find the following particularly relevant:
\citet{Richardson2010-zp} carried out a sociolinguistic study on TV dialogues;
\citet{Silverman1988-dl,Boon2008-rb,OMeara2019-jh} highlights related issues as they pertain to gender and race.

\paragraph{Modeling multi-party conversation structure.} The interaction structure between speakers in a multi-party conversation is shown to be useful for conversation disentanglement~\cite{Mayfield2012-us}. 
More recently, various neural architectures have been proposed to encode utterances and the hierarchical structure of conversations~\cite{Jiang2018-gp,Henderson2020-ir,Wu2020-oq,yu-joty-2020-online}. 
Pre-training with self-supervised tasks \cite{Zhu2020-hx,Wang2020-pz,Gu2021-in} is also used to derive contextual embeddings while factoring in the conversational structure (replacing, for example, next sentence prediction with next \textit{utterance} prediction).
With or without self-supervision, additional embedding layers or attention mechanisms have been proposed to encode the information of conversation structure.
\citet{Gu2020-to,Liu2021-uz} incorporated speaker embeddings and \citet{Sang2022-ao} and \citet{ma-etal-2022-structural} emphasize the interaction between speakers through attention.
For our experiments, as a baseline, we start with a simple and intuitive approach, where we train embedding layers based on structural features like the distance between utterances, and concatenate relevant embedding vectors with standard contextual embeddings from pre-trained language models that we can fine tune for this task.

In this work, conversation structure is situated in the specific domain of films and TV shows, spoken by characters in a scene. 
The structural richness of screenplays makes it an interesting textual representation of movies in NLP~\cite{bhat-etal-2021-hierarchical,Chen2022-yt}, and here, we use screenplays to create an annotated dataset to facilitate future work on conversation disentanglement.

\section{Data}

To study conversation disentanglement in drama, we consider 831 titles: 340 movies and 491 TV series,\footnote{The complete list of titles we used can be found in our GitHub repository.}  randomly sampling one scene from each title for annotation. 
Movies are taken from \textsc{Scriptbase-J}~\cite{Gorinski2018-rh}, based on IMSDB,\footnote{\url{https://imsdb.com/}} because of its extensive coverage of genres and temporal span, along with rich metadata that can adequately support NLP research related to movies. 
Since TV dialogues have different linguistic styles from movie dialogues~\cite{Nelmes2010-oi}, we curated a new dataset, \textsc{TVPilots}, using teleplays of pilot episodes made available by TV Writing.\footnote{\url{https://sites.google.com/site/tvwriting/}} 

\begin{figure}[t]%
  \centering
  \includegraphics[width=\columnwidth]{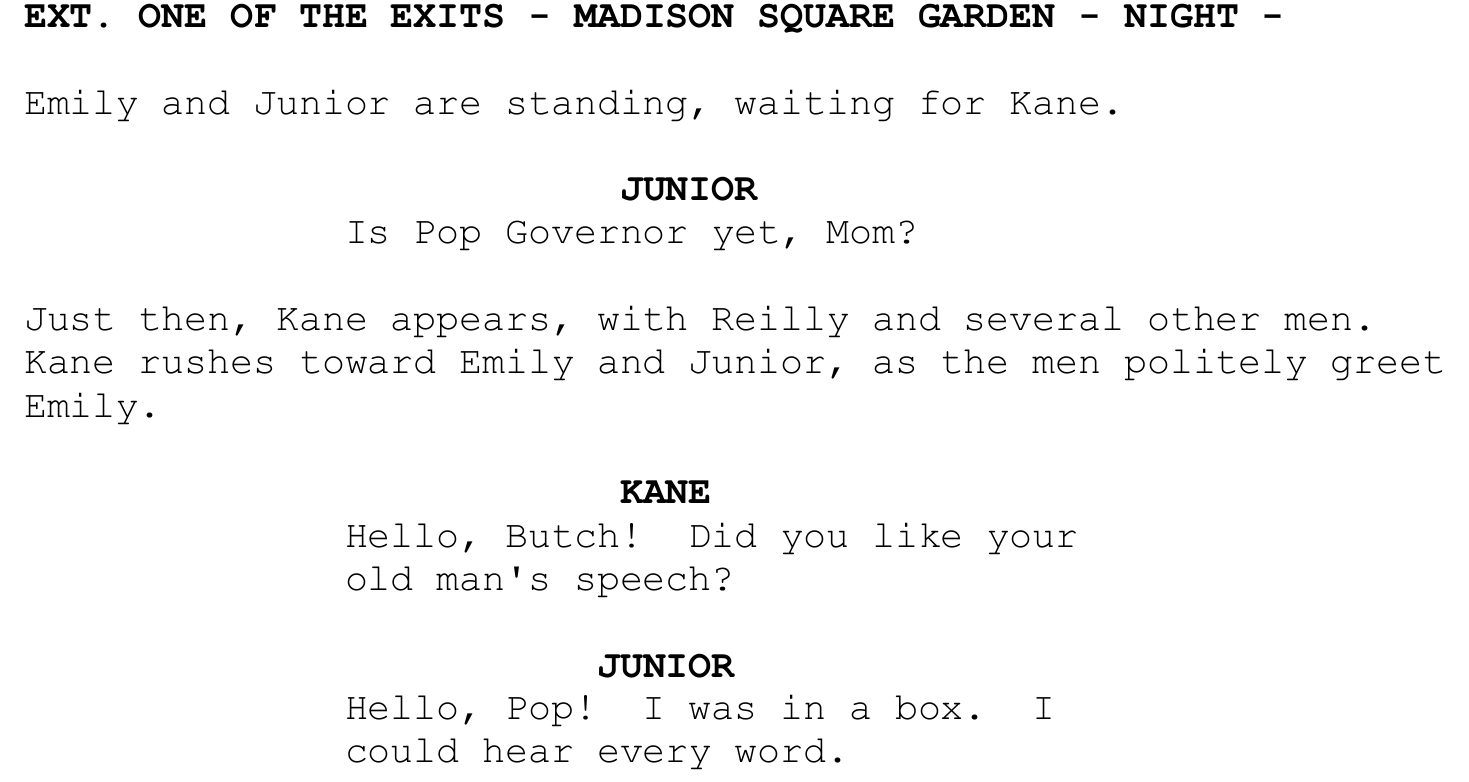}
  \caption{
  Example of the standard screenplay format (\textit{Citizen Kane}). \texttt{EXT. ONE OF THE EXITS} is the scene header. The dialogue portions are usually indented, while action statements (``\textit{Emily and Junior are standing}'') are not indented. Speaker labels are found in cue lines in all caps, followed by their dialogue lines. 
  }
  \label{fig:screenplay-format}
\end{figure}

All screenplays and teleplays from these sources come in the standard format (Fig.~\ref{fig:screenplay-format}):
They have distinct scene headers, speaker labels, and other typographical features for annotators to distinguish between an action statement and a dialogue line.
Since those features are consistent, we have reliable scene markers and speaker labels, and annotators can refer back to the original screenplays when necessary. 
Action statements also helped annotators better understand the scene.

\section{Disentangling drama}\label{sec:anno}

\subsection{Task definition}

We consider conversation disentanglement applied to the domain of scripted conversations in TV series and movies.
On the highest level, given a segmented scene in a screenplay, we want to identify \textit{threads} in a conversation between multiple characters. 
Intuitively, in a scene, a character can change the subject or redirect other characters' attention to themselves, which, in our formulation, means they start a new conversational thread.
Threading drama can help us understand conversational patterns, such as 
who gets to start a new conversation, who dominates the conversation, and how long an average conversation lasts.

The interaction structure involves an \textit{utterance of interest} (UOI) and its \textit{parent utterance}.
An utterance is a dialogue line spoken by a character.
In this work, an utterance of interest holds a directed edge to its parent utterance:
 \begin{equation}
     u_{\textrm{\textsubscript{interest}}} \rightarrow u_\textrm{\textsubscript{parent}} 
 \end{equation}
\noindent Each utterance of interest has only one parent, but one parent can have multiple children. A thread is the transitive closure of all such pairwise links.

\subsection{Threading dramatic conversations}\label{dramatic-conv}

In defining conversational threads in drama, we first adapt~\citeauthor{Goffman1963-ze}'s (\citeyear{Goffman1963-ze}, p. 24) definition of \textit{conversation}:\footnote{We note that thread in other work might be called \textit{sub-conversation}, or simply \textit{conversation}, as opposed to a stream of messages described in e.g.,~\citealp{elsner-charniak-2008-talking}. 
Given the complexity of drama, we chose not to overload the term of \textit{conversation} and preferred \textit{thread}, which also captures their interleaving nature.} a thread is a kind of \textit{focused interaction}: one where ``persons gather close together and openly cooperate to sustain a single focus of attention, typically by taking turns at talking.''
Second, we assume conversations in a given scene in drama are \textit{entangled} and there is often more than one \textit{thread} in any given conversation. 
Taken together, we define
a thread to be a cluster of semantically and pragmatically coherent utterances that are part of a conversation. 
Those utterances share a single, sustainable focus of attention~\cite[cf.~][]{Ervin-Tripp1964-dq,Sacks1974-cg}, either on a character (who has other characters' attention) or a topic (often related to the wants and needs of a character), as well as other observable contextual relations~\cite{ONeill2003-bf}.\footnote{In our annotation framework, threads can cross each other or be resumed later, but we note only ~$0.007\%$ of utterances we annotated exhibit this phenomenon.}

In a conversation, attention can be paid to a character (who has the floor) or a topic (why they are having this conversation): 

\paragraph{Floor.}\citet[p. 392]{Elsner2010-wi} describe the start of a new conversational thread as the process of participants (or in our case, characters) ``hav[ing] refocused their attention \dots~away from whoever held the floor in the parent conversation.'' Like in \citet{Goffman1963-ze}, \textit{attention} is key: the character holding the floor can safely assume that they have attention from others. Such attention is singular and must be sustained throughout the thread; or someone else has the floor.
Sheldon's line off screen in~Fig.~\ref{fig:intro-example} further demonstrates how characters can start a thread by gaining the floor from other characters present in the scene, so they can express their need in their own voice. 
His line, coming out of nowhere, also does not respond to the speech acts of others, making it a typical example of thread starters in drama.

\paragraph{Topic.} Since we follow~\citet{McKee2016-fo} and see dialogues as speech acts, we can often relate the topic of a thread to the desire or intent of the character who started the conversation: 
characters can express their own needs, or respond to someone else's needs, and the need acts as the driving passion of the dramatic conversation.
Operationally, a topic change within a scene usually occurs when the conversation is no longer about the original speech act that starts the scene.
In~Fig.~\ref{fig:intro-example}, Georgie's line, ``Can I drive in with you?'' is an example of topic change and a new expressed intent from a different character in the scene.

For further details and examples, see Appendix~\ref{sec:annotation-guidelines}.

\subsection{Annotation process}

Prior to annotation, all the plays in \textsc{TVPilots} are OCR'd, and all plays were pre-processed following~\citet{bhat-etal-2021-hierarchical}: we extracted the content of structural components (scene headers, character cue lines, dialogue and action lines) in each screenplay and store this information as tabular data.
We further segmented each dialogue line into sentences with spaCy 3.3.0~\cite{honnibal-johnson-2015-improved}, which allows us to annotate at a greater granularity, since sentences in the same dialogue turn can reply to different previous sentences.
We also assigned each scene, action line, dialogue turn, and each sentence in a turn an ID.

\begin{table}
    \centering
    \setlength{\tabcolsep}{4pt}
        \resizebox{0.8\columnwidth}{!}{%
    \begin{tabular}{llcrrr}
        \toprule
        & Metric & Agreement \\
        \midrule
        Link & Pairwise exact match acc. & 92.81 \\
        Cluster & Adjusted Random Index & 79.71 \\
        & $1-$VI & 94.45 \\
        & Shen F\textsubscript{1} & 84.52 \\
        & One-to-One & 81.46 \\
        & Exact match F\textsubscript{1} & 53.33 \\
        \bottomrule
    \end{tabular}
    }
    \caption{\label{tab:agreement}
      Inter-annotator agreement.
    }
\end{table}

All annotations were carried out by the authors of the paper in the span of six months. 
We spent two months on pilot runs as we revised annotation guidelines and discussed edge cases, and the rest on independent annotating.
On average, it took an hour to work through 500 dialogue and action lines.
Our agreement rate is reported in Table~\ref{tab:agreement}; 
we considered standard metrics (described in Appendix~\ref{sec:metrics}) based on 3,271 jointly annotated lines.
Our agreement rate is comparable to previous work~\cite{kummerfeld-etal-2019-large}. 

\section{Experiments}

We compare seven models to see how well existing model architectures that have been proposed for conversation disentanglement perform in the domain of dramatic texts.
We would also like to know if models trained on \citeposs{kummerfeld-etal-2019-large} data, but evaluated on our dramatic data, can leverage its size (seven times larger than ours) to compensate for the striking difference in domain.

\subsection{Notation}
We define a dataset $\mathcal{D} = \{(c^{\mathcal{S}_{i,j}}_{j}, u^{-}_{j}, u^{+}_{j}, u_{i})\}_{i=1}^{|\mathcal{D}|}$ where: 
\begin{itemize}
\itemsep0em 
\item $i$ is the index of an UOI, $j$ that of a candidate parent utterance; an utterance $u$ is a sentence of a dialogue line
\item $\mathcal{S}_{i,j}$ denotes the scene both $u_i$ and $u_j$ are in
\item $c_j$ is the context of $u_j$, defined as all the dialogue and action lines preceding $u_j$ in $\mathcal{S}_{i, j}$
\item $u^+_j$ is a true parent, and $u^-_j$ is a negative example ($u_j \in\mathcal{S}_{i,j}$)
\item $u_i = \{t_i, k_i, w_i^1, w_i^2, \ldots, w_i^m\}$ is an utterance of $m$ tokens spoken by character $k$ in turn $t$; turn information is often given in the play parenthetically as \verb+(CONT'D)+
\end{itemize}

\subsection{Models}

We consider the following models:

\paragraph{Previous.} Adapting~\citet{kummerfeld-etal-2019-large}, we connect all UOIs to their immediate previous utterances; i.e., for $u_i$, its parent utterance is $u_{i-1}$.

\paragraph{Featurized.} \citet{Zhu2021-yq} showed that manually selected features could offer a robust baseline; %
we take inspiration from~\citet{kummerfeld-etal-2019-large} and selected 8 features to train a featurized model: 

{\begin{itemize}%
\itemsep0em 
    \item each utterance: %
    \begin{enumerate} 
        \itemsep0em 
        \item The number of other speakers this character speaks after
        \item The number of utterances ago this character last spoke
        \item Whether the next utterance is spoken by the same character%
    \end{enumerate}
    \item pairwise: between $u_i$ and candidate parent utterance $u_j$ %
    \begin{enumerate}
        \itemsep0em 
        \item[4.] The number of WordPiece tokens $u_i$ and $u_j$ have in common
        \item[5.] The distance between the two utterances $|i-j|$
        \item[6.] Whether there are utterances from either speaker between $u_i$ and $u_j$%
        \item[7.] Whether $u_i$ and $u_j$ are in the same turn%
        \item[8.] Whether $u_i$ and $u_j$ are from the same speaker%
    \end{enumerate}
\end{itemize}}

\paragraph{BERT baseline.}\label{sec:baseline} We adapt the Siamese encoders used in previous work on conversation analysis~\cite{Jiang2018-gp,Henderson2020-ir,Wu2020-oq} to independently encode representations of the utterance of interest, parent utterance, and their associated scene context. 
We used two classes of embeddings: contextual and feature-based.

\textit{Contextual embeddings.} Given a pre-trained model $F$ like BERT, each utterance $u$ spoken by speaker $k$ is stringed together with special tokens as \texttt{[CLS]} $k$ \texttt{[SEP]} $u$ \texttt{[LINE]}, where \texttt{[SEP]} separates a speaker label and the associated line, and \texttt{[LINE]} marks the end of the line.
Here, \texttt{[LINE]} is a custom token whose representation is learned during training.
The contextual embeddings are derived by 
\begin{equation}
    \vect{e}=F(\texttt{[CLS]} k \texttt{[SEP]} u \texttt{[LINE]})
\end{equation}
We extract the~\verb+[CLS]+~token, denoted $\vect{e}^{\texttt{[CLS]}}$. 

\textit{Feature-based embeddings.} To enhance the expressivity of our models, we introduced additional embedding layers, randomly initialized, to encode information pertinent to the conversation structure in the scene. 
Each of the following features is assigned an embedding vector: 
utterance distance $\vect{f}_{d_{i}}$ (feature 5 from the featurized model), turn $\vect{f}_{t_{i}}$ (whether this line in the same turn as last, feature 7), scene speaker $\vect{f}_{k_{i}}$ (whether two speakers are the same, feature 8). 
All $\vect{f}\in \mathbbm{R}^{250\times2}$. 
We can then represent each utterance pair as the concatenation of all embeddings:
\begin{equation}
    \begin{aligned}
        \vect{u}_{i,j} = [&\vect{e}_{c_j}^{\texttt{[CLS]}}; \vect{e}_{u_j}^{\texttt{[CLS]}}; \vect{e}_{u_i}^{\texttt{[CLS]}}; \\ 
        & \vect{e}_{u_i}^{\texttt{[CLS]}} - \vect{e}_{u_j}^{\texttt{[CLS]}}; \vect{e}_{u_i}^{\texttt{[CLS]}} \odot \vect{e}_{u_j}^{\texttt{[CLS]}}; \\
        & \vect{f}_{k_{i}}; \vect{f}_{t_{i}}; \vect{f}_{d_{i,j}}]
    \end{aligned}
\end{equation}
\noindent Finally, we pass $\vect{u}_{i,j}$ through a non-linearity before the sigmoid output layer to compute the matching score $m_{i, j}$:
\begin{align}
    \vect{h}_{i,j}=&\tanh(\vect{w}^{(0)\top}\vect{u}_{i,j}) \label{eq:m1} \\
    m_{i,j}=&\sigma(\vect{w}^{(1)\top}\vect{h}_{i,j}) \label{eq:m2} 
\end{align}

For training, we used a self-link token \verb+[SELF]+ as a parent candidate for every $u_i$, which is assigned as the true parent if $u_i$ is the start of a thread. 
For each $(u_j^+, u_i)$ pair in our annotation, we sample five $u_j^-$.
The objective is to minimize the binary cross-entropy loss, $\mathcal{L}_{\text{link}}$. 

At inference time, for each $u_i$ we have a candidate pool $\mathcal{P}_i=\{u_{i}, u_{i-1}, \dots, u_{i-C-1}\}$ to consider $u_i$ itself (self-pointing) along with $C-1$ previous candidates. 
Since in 90\% of our annotation the true parent is within 5 utterances, we picked candidate pool size $C=6$. 
We calculate the matching score between $u_i$ and all $u_c \in \mathcal{P}_i$ and select $\mathrm{argmax}_{\mathcal{P}_i}$~$m_{i, c}$ as parent.

In addition to the BERT baseline, we adapt three recent architectures for dramatic conversation disentanglement.
They are designed with \citeposs{kummerfeld-etal-2019-large} IRC chat log in mind, and while many textual features do not have equivalents in our dramatic domain, we incorporate some designs as we saw fit, described below:

\paragraph{BERT with soft attention alignment.} We adapt the soft alignment mechanism in the pointer module from~\citet{yu-joty-2020-online} to emphasize the textual similarity between $u_i$ and $u_j$:
\begin{eqnarray}
        \vect{H}_i^{'}=&\mathrm{softmax}(\vect{H}_i \vect{H}_j^\top) \vect{H}_j  \label{eq:qhi}  \\
        \vect{H}_j^{'}=&\mathrm{softmax}(\vect{H}_j \vect{H}_i^\top) \vect{H}_i  \\
    \vect{h}_i^{f}=&[\vect{h}_i; \vect{h}_i^{'}; \vect{h}_i - \vect{h}_i^{'}; \vect{h}_i \odot \vect{h}_i^{'}]\label{eq:qhj1} \\
    \vect{h}_j^{f}=&[\vect{h}_j; \vect{h}_j^{'}; \vect{h}_j - \vect{h}_j^{'}; \vect{h}_j \odot \vect{h}_j^{'}]\label{eq:qhj2} 
\label{eq:qhj}
\end{eqnarray}

\noindent where $\vect{H}_i = (\vect{h}_{i,0}, \ldots, \vect{h}_{i,p})$ and $\vect{H}_j = (\vect{h}_{j,0}, \ldots, \vect{h}_{j,q})$ are the bidrectional LSTM representations for $u_i$ and $u_j$. 
$\vect{H}_i$ is used as query vectors to compute attentions over the key/value vectors in $\vect{H}_j$ and the set of attended vectors $\vect{H}_i^{'} $, one for each $\vect{h}_i \in \vect{H}_i$. 
In Eq. \ref{eq:qhj1}--\ref{eq:qhj2} we enhance the interactions by applying difference and element-wise product between the original and attended vectors. 
Finally, we swap out BERT-based contextual embeddings for $u_i$ and $u_j$ with $\vect{h}_i^{f}$ and $\vect{h}_j^{f}$, with the following resultant representation of the two utterances:
\begin{equation}
    \begin{aligned}
        \vect{u}_{i,j} = [&\vect{e}_{c_j}^{\texttt{[CLS]}}; \vect{h}_i^{f}; \vect{h}_j^{f}; \\
        & \vect{f}_{k_{i}}; \vect{f}_{t_{i}}; \vect{f}_{d_{i,j}}]%
    \end{aligned}
\end{equation}
\noindent The matching score is calculated using Eq.~\ref{eq:m1}--\ref{eq:m2}.

\paragraph{6-way classifier.} The structural characterization of conversation proposed by~\citet{ma-etal-2022-structural} is the current state of the art.  
We use their architecture without reference dependency modeling, since we don't have mentions in our movie data in the same format as IRC, but retain the rest.
Their goal is to train a $C$-way classifier: for each, $u_i$, pick one from candidates including $u_i$ and $u_{i-j}$ ($1 \leq j \leq C-1$). 
The UOI and candidate pairs are stringed and fed into a pre-trained model as 
\begin{equation}
    \vect{H}_0=F(\texttt{[CLS]} u_{i-j} \texttt{[SEP]} u_i), 
\end{equation}
$\vect{H}_0\in\mathbbm{R}^{C\times L\times|F|}$, where $L$ is the input sequence length.
To obtain aggregated contextualized representations, we extract the~\texttt{[CLS]}~token: $\vect{H}_1 =\vect{H}_0^{\texttt{[CLS]}}$, $\vect{H}_1\in\mathbbm{R}^{C\times|F|}$.
For candidate window size, we chose $C=6$ (including self-pointing as one candidate).

Their architecture features two components: speaker property modeling and the Syn-LSTM module. 
Speaker property modeling leverages the masked Multi-Head Self-Attention (MHSA, \citealp{Liu2021-uz}) mechanism to account for utterances from the same speaker with a speaker-aware mask matrix $M$, which we include in our adaptation:
\begin{equation}
\begin{split}
M[i, j] &=
\begin{cases}
0,& k_i = k_j \\
-\infty,& \text{otherwise}\\
\end{cases}\\
\end{split}
\label{eq2}
\end{equation}
\noindent Syn-LSTM~\cite{xu-etal-2021-better} is a biLSTM with an additional input gate to retain the information of utterances within the candidate window, designed to make the model context-aware. 
In other words, we have $\vect{H}_2 = \text{MHSA}(\vect{H}_1, M)$ and $\vect{H}_3 = \text{Syn-LSTM}(\vect{H}_2)$, where $\vect{H}_2, \vect{H}_3 \in \mathbbm{R}^{C\times|F|}$. In this structural characterization, the final representation between each $[u_i,u_{i-j}]$ pair ($1 \leq j \leq C-1$) and the self-pointing $[u_i,u_i]$ pair is:
\begin{equation}
\vect{h}_{i, j} = \vect{[p}_{ii}, \vect{p}_{ij}, \vect{p}_{ii} \odot \vect{p}_{ij}, \vect{p}_{ii} - \vect{p}_{ij}\vect{]}, 
\end{equation}
where $\vect{p}_{ij}$ is the representation for the pair of $[u_i,u_{i-j}]$ from $\vect{H}_3$. 
$\vect{h}_{i, j}$ is then fed into the classification head to predict the parent. The training objective is to minimize the cross-entropy loss.

\paragraph{Multi-task learning.} We follow~\citet{yu-joty-2020-online,Zhu2021-yq,Huang2022-re} and introduce an auxiliary task, a binary classifier to predict whether $u_i$ and $u_j$ belong to the same thread, the probability of which is:
\begin{eqnarray}
    t_{i, j} = \sigma(\vect{w}_t^\top\vect{u}_{i,j}) 
\end{eqnarray}
\noindent where $\vect{u}_{i,j}$ is the representation of the utterance pair given.
The objective is to minimize the binary cross-entropy loss: 
\begin{equation}
\mathcal{L}_{\text{thread}}= - y_{i,j} \log t_{i, j} - (1 - y_{i,j}) \log (1-t_{i, j})
\end{equation}
where $y_{i,j} = 1$ if $u_i$ and  $u_j$ are in the same thread, $0$ otherwise. The total training loss $\mathcal{L}$ of this model class is: 
\begin{equation}
\mathcal{L} = \mathcal{L}_{\text{link}} + \alpha\mathcal{L}_{\text{thread}}
\end{equation}
\noindent where $\alpha$ is a hyper-parameter to control the impact of the auxiliary task. 
We experimented with $\alpha=\{0.1, 0.5, 1.0\}$ and $0.1$ performed best.

\paragraph{BERT baseline trained with~\citeposs{kummerfeld-etal-2019-large} data.} Lastly, to test the influence of domain difference between IRC chat logs and dramatic conversations, we trained our models instead on~\citeposs{kummerfeld-etal-2019-large} training data, to be evaluated on our drama data. 
We extracted all usernames as speaker labels and treated system messages as action statements. 
Since individual users can send multiple messages in what would be one dialogue turn in movies, we did not perform sentence segmentation on messages.

 \begin{table}[!t]
    \centering
    \begin{tabular}{lrrrrrr}
    \toprule
    set &  train & dev & test  \\
    \midrule
    \# titles & 563 & 127 & 141 \\
    \# unique speakers & 1,711 & 371 & 389 \\
    \# dialogue lines & 11,672 & 2,639 & 2,743 \\
    \# turns & 5,988 & 1,298 & 1,475  \\
    \# action lines & 8,756 & 2,059 & 1,980 \\
    \bottomrule
    \end{tabular}%
    \caption{\label{tab:anno_data_stats}
      Data statistics.
    }
\end{table}

\begin{table*}[!htp]
    \centering
    \setlength{\tabcolsep}{4pt}
    \resizebox{1.0\textwidth}{!}{%
    \aboverulesep=0ex %
    \belowrulesep=0ex %
    \begin{tabular}{@{}lr|rrrrr@{}}
        \toprule 
                          & Link prediction & \multicolumn{5}{c}{Clustering}   \\
        Model   & Acc.   & ARI  & $1-$VI  & Shen F\textsubscript{1}  & 1-1 & Exact match F\textsubscript{1} \\
        \midrule 
        \multicolumn{7}{@{}l}{\textbf{trained with~\citeauthor{kummerfeld-etal-2019-large} data}} \\
        \midrule
        BERT baseline & 51.10 {\small[49.28-52.96]}& 48.64 {\small[45.48-51.78]}& 83.67 {\small[82.89-84.43]}& 62.05 {\small[60.22-64.07]}& 54.60 {\small[52.67-56.62]}& 6.42 {\small[4.67-8.13]}\\        
        6-way classifier & 60.84 {\small[59.35-62.35]}& 55.10 {\small[51.16-59.46]}& 86.85 {\small[85.88-87.90]}& 63.65 {\small[60.77-66.97]}& 60.20 {\small[57.20-63.56]}& 11.62 {\small[8.91-14.37]}\\
        \midrule 
        \multicolumn{7}{@{}l}{\textbf{trained with our dataset}} \\
        \midrule
        Previous & 90.26 {\small[89.78-90.75]}& 46.69 {\small[45.30-48.72]}& 85.29 {\small[84.65-86.20]}& 54.80 {\small[53.09-57.31]}& 51.89 {\small[50.13-53.99]}& 14.95 {\small[12.13-17.50]}\\
        Featurized & 89.75 {\small[88.86-90.65]}& 47.80 {\small[44.53-52.28]}& 85.61 {\small[84.54-86.88]}& 55.90 {\small[52.77-59.70]}& 53.05 {\small[49.64-57.13]}& 15.25 {\small[11.66-19.14]}\\
         BERT baseline & 89.44 {\small[88.49-90.44]}& 57.98 {\small[54.17-63.20]}& 88.78 {\small[87.85-89.83]}& 66.55 {\small[63.66-69.89]}& 63.71 {\small[60.55-67.43]}& 25.25 {\small[20.90-29.73]}\\
         \;\; {\footnotesize + attn. alignment} & \textbf{90.28} {\small[89.31-91.27]}& 57.28 {\small[53.35-62.47]}& 88.62 {\small[87.54-89.80]}& 65.37 {\small[62.17-69.04]}& 62.78 {\small[59.23-66.81]}& \textbf{25.88} {\small[21.11-30.63]}\\
         \;\; {\footnotesize + aux. task} & 90.12 {\small[89.26-91.02]}& 53.32 {\small[49.37-59.02]}& 87.63 {\small[86.59-88.86]}& 62.25 {\small[59.12-66.06]}& 59.81 {\small[56.34-63.93]}& 21.20 {\small[17.19-25.54]}\\
         \;\;\;\; {\footnotesize+ both} & 90.24 {\small[89.34-91.17]}& 57.63 {\small[54.04-62.15]}& 88.60 {\small[87.61-89.72]}& 65.54 {\small[62.61-69.01]}& 63.21 {\small[60.03-66.98]}& 25.27 {\small[20.97-30.03]}\\
         6-way classifier & 87.23 {\small[86.24-88.27]}& \textbf{64.81} {\small[60.70-69.98]}& \textbf{90.11} {\small[89.28-91.05]}& \textbf{72.20} {\small[69.63-75.29]}& \textbf{69.02} {\small[66.30-72.19]}& 25.40 {\small[21.67-29.27]}\\
        \bottomrule
    \end{tabular}}
    \caption{\label{tab:experiments}
      Experimental results. All metrics are reported with 95\% bootstrap confidence intervals.
    }
\end{table*}

\subsection{Setup}

We trained all our models for 10 epochs and used the dev set for early stopping with the learning rate $5\times 10^{-6}$ (BERT-based) and $10^{-3}$ (linear).
Our train--test split is reported in Table~\ref{tab:anno_data_stats}.
For our BERT implementation, we used~\texttt{bert-base-cased} from HuggingFace 4.19.2 with PyTorch 1.10.0.\footnote{\url{https://huggingface.co/bert-base-cased}; \url{https://pytorch.org/}.}
An epoch took 1 hour 8 minutes on average on two NVIDIA GeForce RTX 2080 Ti GPUs.

\subsection{Results}
Experimental results are presented in Table~\ref{tab:experiments}. We report the standard set of metrics (described in Appendix~\ref{sec:metrics}), along with their 95\% bootstrap confidence intervals.
We first note that the clustering metrics are low while link prediction accuracy is high because the most reasonable parent utterance for most UOIs (90\%) is the immediately previous utterance, which leads to high baseline accuracy for link prediction, but low clustering baselines when considering entire threads. 

The performance of models trained on \citeposs{kummerfeld-etal-2019-large} suggests that domain difference matters.
While seven times larger, their dataset is in an entirely different domain, and intuitively, chatroom users interact differently from movie characters. 
Such differences might account for the inferior performance, especially on stricter cluster metrics like One-to-One and Exact Match. 

Enhancements to the baseline lead to minor, statistically insignificant, improvements, and the 6-way classifier outperforms the rest model classes on most metrics. 
Therefore, for the analysis below, we use the 6-way classifier.

\section{Analysis}

To illustrate the usefulness of conversation disentanglement in drama, we disentangled 808 movies from \textsc{Scriptbase-J} and carried out two analyses enabled by this work to explore two questions that engage previous work in film studies:\footnote{The list of movies we used for analysis can be found in our GitHub repo: \url{https://github.com/kentchang/dramatic-conversation-disentanglement/blob/bf3d2fbc00f9d64356c308a2c0ca6b2e73580c19/list/titles-for-analysis.txt}}
\paragraph{Are conversational threads in movies getting shorter over the years?} In his analysis on visual style in contemporary films, film historian \citet[p. 16]{Bordwell2002-rn} observed, ``For many of us, today's popular American cinema is always fast'': 
the average shot length is decreasing and cuts and camera movements have become more rapid over the course of the twentieth century, leading to an impression of ``intensified continuity.'' 
A similar observation is made in~\citeposs{Cutting2010-kw} empirical work, which relates this trend to our natural fluctuation of attention. 
Notably, such work emphasizes the visual aspect of films, which reinforces the established hierarchy in film studies: the film is a visual medium, and image is more important than sound.
This hierarchy is critiqued in studies on film dialogues in particular~\cite{Kozloff2000-ox}, since characters converse with one another only after the advent of sound films.
It then leads us to ask:
Are conversational threads in films also getting shorter over the years?

\begin{figure}[!htp]
  \centering
  \includegraphics[width=0.72\columnwidth]{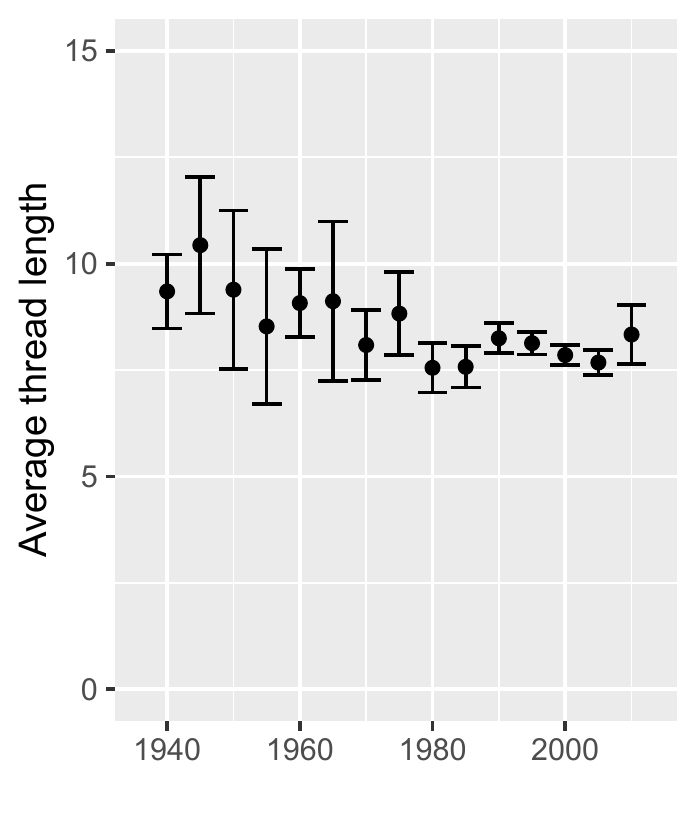}
  \caption{
  The average thread lengths of movies in a 5-year range, along with 95\% confidence intervals.
  }
  \label{fig:avg_threads_5yr}
\end{figure}

Since we have disentangled movie conversations into threads, we can calculate the average number of utterances there are in a thread in a given movie and in a given year. 
Our movie data, while spanning from 1930s to 2010s, is not evenly distributed. As a result, for this analysis, we aggregated movies in a 5-year range. 
While we would expect thread lengths to also decrease, Fig.~\ref{fig:avg_threads_5yr} tells a different story.
We can see the average thread length decreasing (although not statistically significantly so) until 1970, and the trend is flat since.
Film dialogues seem to be resisting the broader trend associated with visual styles.

\paragraph{What is the pattern of floor claiming between men and women in movies?} It has been pointed out that the film industry became dominated increasingly by men over the twentieth century~\cite{Boon2008-rb}. 
The Bechdel Test~\cite{Bechdel1986-uj} is a popular and well-known measure for the representation of women in films, often used for advocating that women on screen should ``speak up''~\cite{OMeara2016-ek} to encourage more diverse representation. %
Through this work, we would like to add an additional dimension to it:
How often do characters who are women start a conversation in films?

In the tradition of continental philosophy, to initiate a conversation---or, to become a \textit{speaking subject}---is a socially and ethically significant act~\cite{Foucault1972-dk,Lacan2006-xg}.
This influenced much of feminist film studies that considers the presence and absence of women's voices in films~\cite{Silverman1988-dl,Lawrence1991-lk,Sjogren2006-rj}.
This work inspires us to frame and measure the agency of characters who are women in relation to the frequency with which they get to start a new conversational thread and claim the floor. 

\begin{figure}[!htp]
  \centering
  \includegraphics[width=0.95\columnwidth]{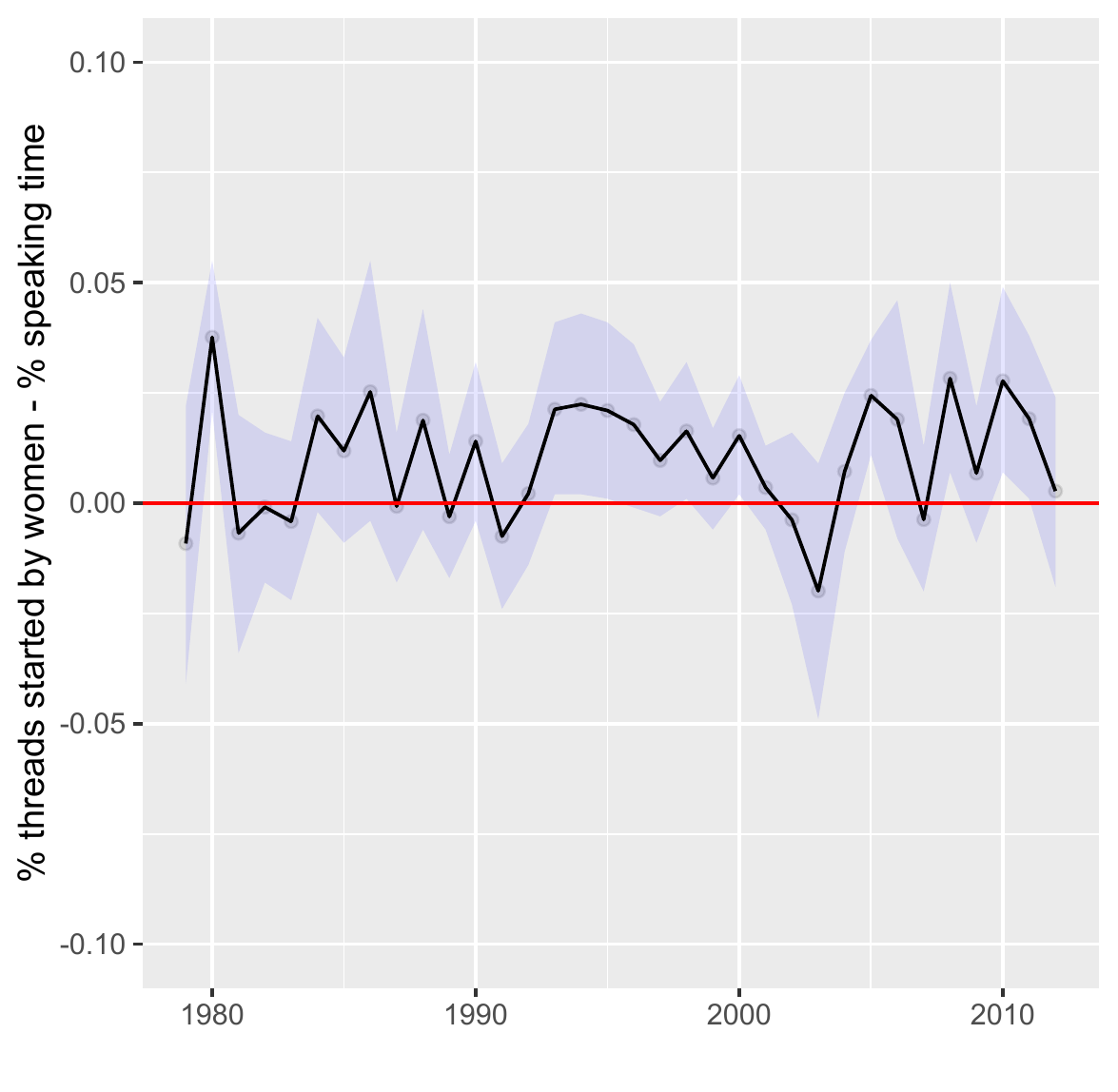}
  \caption{
  The percentage of threads started by women relative to their speaking time, along with 95\% CIs.
  }
  \label{fig:startMinusSpeaking}
\end{figure}

We start by using TMDb's API to look up the gender of the actor portraying the character.\footnote{\url{https://developers.themoviedb.org/3/people/get-person-details}}
For this analysis, we only consider movies released after 1979, after which point we have at least five movies each year.%
In our data, $30.4\%$ of threads are started by women.
This is aligned with the oft-stated observation that men talk more than women in films~\cite{ramakrishna-etal-2017-linguistic,Lauzen2019-ta}, and the trend has not significantly changed for decades. 

However, as we see in Fig.~\ref{fig:startMinusSpeaking}, when we subtract the percentage of \textit{speaking time} by women (the number of lines they have) from that of \textit{threads started} by women (e.g., in 2011, $32.7\%$ of threads were started by women, and $30.8\%$ of lines were spoken by women, so we see an absolute difference of $+1.9$), we see that women generally start more threads relative to their speaking time, and that is also relatively constant over time.
In the figure, any year in which the 95\% confidence interval does not overlap with 0 is significant at that level; while this is not significant for many individual years (given the limited number of movies per year), it is significant over all years ($+1.0, [0.07, 0.14]$).
This finding is surprising because it suggests that despite their under-representation, women characters are written to initiate conversations more than their male counterparts.

\section{Conclusion}

We present in this work a new dataset for studying conversation disentanglement in movies and TV shows in order to enrich the landscape of this line of work in NLP.
Movie and TV dialogues offer pragmatic patterns and interaction structures different from chat logs, on which standard benchmarks for this task are built.
To ensure high quality of this dataset, we digitized teleplays written for TV pilots, so we have more screenplays in the standard format, which we find most useful for annotation. 
In addition, we draw on theoretical resources from sociolinguistics, sociology, and film studies to create a robust annotation scheme that considers topic and floor changes specifically in the context of drama, which we believe speaks to the needs of the wider scholarly community. 

To the best of our knowledge, no previous technical or theoretical work has offered a working operationalization of conversational thread in the context of dramatic (scripted, spoken, face-to-face) conversations, or examined the significance of initiating a conversation or gaining the floor in this domain.
While we do not claim that the results from our analysis are definitive, our work has demonstrated a new method to further investigate the sound-image hierarchy, gendered power dynamics in films, and communication behaviors in cultural representations on screen. 
We hope this will encourage and facilitate future research on drama and conversation in NLP, film studies, and the computational humanities.

\section*{Acknowledgments}

We thank Timea Ryan for feedback on our annotation guidelines, and our anonymous reviewers for their constructive comments.
The research reported in this article was supported by funding from the National Science Foundation (IIS-1942591) and the National Endowment for the Humanities (HAA-271654-20).

\section*{Limitations}

This work is first limited by the availability of screenplays in the standard format. 
While movie or TV show transcripts are more readily available (subject to permission to use), they are less ideal for annotation due to unreliable scene headers and speaker labels. 
This therefore limits the size of our corpus, as digitization and correction are labor-intensive.
In our analysis, we relied on metadata from \textsc{Scriptbase-J} (each movie has a Jinni\footnote{\url{https://en.wikipedia.org/wiki/Jinni_(search_engine)}} profile that includes its corresponding IMDb\footnote{\url{https://www.imdb.com/}} ID) and TMDb.\footnote{\url{https://www.themoviedb.org/}}
Given it scale, we weren't able to check individually whether the release year or the gender of actors in those community-built resources is correct or up-to-date.
This work is also unimodal, while movies and TV shows are multimodal, which meant we did not have access to the video for annotation, and we could not compare thread length and shot length, among other things.

\section*{Ethics Statement}

We are aware that this dataset and the analytical work that follows only represent a limited set of cultural and ethnic groups as well as language uses.
The dataset we're annotating highlights US movies (and not e.g., Bollywood, Nollywood or the global film industry more generally), and so one risk is the centering that culture (and conversational norms) within that dataset at the expense of others.
There have been documented allocational and distributional biases in the film industry~\cite{Baker1991-kf,Ravid1999-ji,OBrien2014-qv,Khadilkar2022-pl}, and we encourage those interested in furthering this line of work to acquaint themselves with relevant discourses.
We are also aware that the dataset contains potentially problematic content, such as vulgar, violent, or offensive language in screenplays, or other biases held by individual screenwriters.

\bibliography{anthology,custom,dbcustom}

\begin{thebibliography}{66}
\expandafter\ifx\csname natexlab\endcsname\relax\def\natexlab#1{#1}\fi

\bibitem[{Azab et~al.(2019)Azab, Kojima, Deng, and
  Mihalcea}]{azab-etal-2019-representing}
Mahmoud Azab, Noriyuki Kojima, Jia Deng, and Rada Mihalcea. 2019.
\newblock \href {https://doi.org/10.18653/v1/K19-1010} {Representing movie
  characters in dialogues}.
\newblock In \emph{Proceedings of the 23rd Conference on Computational Natural
  Language Learning (CoNLL)}, pages 99--109, Hong Kong, China. Association for
  Computational Linguistics.

\bibitem[{Baker and Faulkner(1991)}]{Baker1991-kf}
Wayne~E Baker and Robert~R Faulkner. 1991.
\newblock \href {https://doi.org/10.1086/229780} {{Role as Resource in the
  Hollywood Film Industry}}.
\newblock \emph{The American journal of sociology}, 97(2):279--309.

\bibitem[{Bechdel(1986)}]{Bechdel1986-uj}
Alison Bechdel. 1986.
\newblock \href {https://play.google.com/store/books/details?id=f9XHQgAACAAJ}
  {\emph{{Dykes to Watch Out for}}}.
\newblock Firebrand Books.

\bibitem[{Bhat et~al.(2021)Bhat, Saluja, Dye, and
  Florjanczyk}]{bhat-etal-2021-hierarchical}
Gayatri Bhat, Avneesh Saluja, Melody Dye, and Jan Florjanczyk. 2021.
\newblock \href {https://doi.org/10.18653/v1/2021.nuse-1.1} {Hierarchical
  encoders for modeling and interpreting screenplays}.
\newblock In \emph{Proceedings of the Third Workshop on Narrative
  Understanding}, pages 1--12, Virtual. Association for Computational
  Linguistics.

\bibitem[{Boon(2008)}]{Boon2008-rb}
Kevin~Alexander Boon. 2008.
\newblock \href {https://play.google.com/store/books/details?id=DY9VEZIDDgMC}
  {\emph{{Script Culture and the American Screenplay}}}.
\newblock Wayne State University Press.

\bibitem[{Bordwell(2002)}]{Bordwell2002-rn}
David Bordwell. 2002.
\newblock \href {https://online.ucpress.edu/fq/article-abstract/55/3/16/28821}
  {{Intensified continuity visual style in contemporary American film}}.
\newblock \emph{Film quarterly}, 55(3):16--28.

\bibitem[{Chen et~al.(2022)Chen, Chu, Wiseman, and Gimpel}]{Chen2022-yt}
Mingda Chen, Zewei Chu, Sam Wiseman, and Kevin Gimpel. 2022.
\newblock \href {https://aclanthology.org/2022.acl-long.589} {{{S}umm{S}creen:
  A Dataset for Abstractive Screenplay Summarization}}.
\newblock In \emph{{Proceedings of the 60th Annual Meeting of the Association
  for Computational Linguistics (Volume 1: Long Papers)}}, pages 8602--8615,
  Dublin, Ireland. Association for Computational Linguistics.

\bibitem[{Cutting et~al.(2010)Cutting, DeLong, and Nothelfer}]{Cutting2010-kw}
James~E Cutting, Jordan~E DeLong, and Christine~E Nothelfer. 2010.
\newblock \href {http://dx.doi.org/10.1177/0956797610361679} {{Attention and
  the evolution of Hollywood film}}.
\newblock \emph{Psychological science}, 21(3):432--439.

\bibitem[{Danescu-Niculescu-Mizil and
  Lee(2011)}]{danescu-niculescu-mizil-lee-2011-chameleons}
Cristian Danescu-Niculescu-Mizil and Lillian Lee. 2011.
\newblock \href {https://aclanthology.org/W11-0609} {Chameleons in imagined
  conversations: A new approach to understanding coordination of linguistic
  style in dialogs}.
\newblock In \emph{Proceedings of the 2nd Workshop on Cognitive Modeling and
  Computational Linguistics}, pages 76--87, Portland, Oregon, USA. Association
  for Computational Linguistics.

\bibitem[{Elsner and Charniak(2008)}]{elsner-charniak-2008-talking}
Micha Elsner and Eugene Charniak. 2008.
\newblock \href {https://aclanthology.org/P08-1095} {You talking to me? a
  corpus and algorithm for conversation disentanglement}.
\newblock In \emph{Proceedings of ACL-08: HLT}, pages 834--842, Columbus, Ohio.
  Association for Computational Linguistics.

\bibitem[{Elsner and Charniak(2010)}]{Elsner2010-wi}
Micha Elsner and Eugene Charniak. 2010.
\newblock \href {https://aclanthology.org/J10-3004} {{Disentangling Chat}}.
\newblock \emph{Computational Linguistics}, 36(3):389--409.

\bibitem[{Ervin-Tripp(1964)}]{Ervin-Tripp1964-dq}
Susan Ervin-Tripp. 1964.
\newblock \href {http://doi.wiley.com/10.1525/aa.1964.66.suppl_3.02a00050} {{An
  analysis of the interaction of language, topic, and listener}}.
\newblock \emph{American anthropologist}, 66(6\_PART2):86--102.

\bibitem[{Foucault(1972)}]{Foucault1972-dk}
Michel Foucault. 1972.
\newblock \emph{{The Archaeology of Knowledge}}.
\newblock Pantheon Books, New York.

\bibitem[{Goffman(1963)}]{Goffman1963-ze}
Erving Goffman. 1963.
\newblock \href {https://play.google.com/store/books/details?id=EM1NNzcR-V0C}
  {\emph{{Behavior in Public Places}}}.
\newblock The Free Press, New York.

\bibitem[{Goodwin(1981)}]{Goodwin1981-wq}
Charles Goodwin. 1981.
\newblock \emph{{Conversational organization: Interaction between speakers and
  hearers}}.
\newblock Academic Press.

\bibitem[{Gorinski and Lapata(2015)}]{gorinski-lapata-2015-movie}
Philip~John Gorinski and Mirella Lapata. 2015.
\newblock \href {https://doi.org/10.3115/v1/N15-1113} {Movie script
  summarization as graph-based scene extraction}.
\newblock In \emph{Proceedings of the 2015 Conference of the North {A}merican
  Chapter of the Association for Computational Linguistics: Human Language
  Technologies}, pages 1066--1076, Denver, Colorado. Association for
  Computational Linguistics.

\bibitem[{Gorinski and Lapata(2018)}]{Gorinski2018-rh}
Philip~John Gorinski and Mirella Lapata. 2018.
\newblock \href {https://aclanthology.org/N18-1160} {{What's This Movie About?
  A Joint Neural Network Architecture for Movie Content Analysis}}.
\newblock In \emph{{Proceedings of the 2018 Conference of the North American
  Chapter of the Association for Computational Linguistics: Human Language
  Technologies, Volume 1 (Long Papers)}}, pages 1770--1781, New Orleans,
  Louisiana. Association for Computational Linguistics.

\bibitem[{Gu et~al.(2020)Gu, Li, Liu, Ling, Su, Wei, and Zhu}]{Gu2020-to}
Jia-Chen Gu, Tianda Li, Quan Liu, Zhen-Hua Ling, Zhiming Su, Si~Wei, and
  Xiaodan Zhu. 2020.
\newblock \href {https://dl.acm.org/doi/10.1145/3340531.3412330}
  {{Speaker-Aware BERT for Multi-Turn Response Selection in Retrieval-Based
  Chatbots}}.
\newblock In \emph{{Proceedings of the 29th ACM International Conference on
  Information \& Knowledge Management}}.

\bibitem[{Gu et~al.(2022)Gu, Tao, and Ling}]{Gu2022-hj}
Jia-Chen Gu, Chongyang Tao, and Zhen-Hua Ling. 2022.
\newblock \href {https://doi.org/10.24963/ijcai.2022/768} {{Who says what to
  whom: A survey of multi-party conversations}}.
\newblock In \emph{{Proceedings of the Thirty-First International Joint
  Conference on Artificial Intelligence}}, California. International Joint
  Conferences on Artificial Intelligence Organization.

\bibitem[{Gu et~al.(2021)Gu, Tao, Ling, Xu, Geng, and Jiang}]{Gu2021-in}
Jia-Chen Gu, Chongyang Tao, Zhenhua Ling, Can Xu, Xiubo Geng, and Daxin Jiang.
  2021.
\newblock \href {https://aclanthology.org/2021.acl-long.285} {{MPC-BERT: A
  Pre-Trained Language Model for Multi-Party Conversation Understanding}}.
\newblock In \emph{{Proceedings of the 59th Annual Meeting of the Association
  for Computational Linguistics and the 11th International Joint Conference on
  Natural Language Processing (Volume 1: Long Papers)}}, pages 3682--3692,
  Online. Association for Computational Linguistics.

\bibitem[{Halkidi et~al.(2002)Halkidi, Batistakis, and
  Vazirgiannis}]{Halkidi2002-rl}
Maria Halkidi, Yannis Batistakis, and Michalis Vazirgiannis. 2002.
\newblock {Cluster validity methods: part I}.
\newblock \emph{ACM Sigmod Record}.

\bibitem[{He and Herman(1998)}]{He1998-dk}
Agnes~Weiyun He and Vimala Herman. 1998.
\newblock \href
  {https://www.taylorfrancis.com/books/mono/10.4324/9780203981108/dramatic-discourse-vimala-herman}
  {{Dramatic discourse: Dialogue as interaction in plays}}.
\newblock \emph{Language}, 74(2):384.

\bibitem[{Heldman(2016)}]{Heldman2016}
Caroline Heldman. 2016.
\newblock Hitting the bullseye: Reel girl archers inspire real girl archers.
\newblock Technical report, Geena Davis Institute.

\bibitem[{Henderson et~al.(2020)Henderson, Casanueva, Mrk{\v s}i{\'c}, Su, Wen,
  and Vuli{\'c}}]{Henderson2020-ir}
Matthew Henderson, I{\~n}igo Casanueva, Nikola Mrk{\v s}i{\'c}, Pei-Hao Su,
  Tsung-Hsien Wen, and Ivan Vuli{\'c}. 2020.
\newblock \href {https://aclanthology.org/2020.findings-emnlp.196}
  {{{C}onve{RT}: Efficient and Accurate Conversational Representations from
  Transformers}}.
\newblock In \emph{{Findings of the Association for Computational Linguistics:
  EMNLP 2020}}, pages 2161--2174, Online. Association for Computational
  Linguistics.

\bibitem[{Honnibal and Johnson(2015)}]{honnibal-johnson-2015-improved}
Matthew Honnibal and Mark Johnson. 2015.
\newblock \href {https://doi.org/10.18653/v1/D15-1162} {An improved
  non-monotonic transition system for dependency parsing}.
\newblock In \emph{Proceedings of the 2015 Conference on Empirical Methods in
  Natural Language Processing}, pages 1373--1378, Lisbon, Portugal. Association
  for Computational Linguistics.

\bibitem[{{hooks}(1992)}]{hooks92}
{bell} {hooks}. 1992.
\newblock \emph{Black Looks: Race and Representation}.
\newblock South End Press.

\bibitem[{Huang et~al.(2022)Huang, Zhang, Fei, and Liao}]{Huang2022-re}
Chengyu Huang, Zheng Zhang, Hao Fei, and Lizi Liao. 2022.
\newblock \href {https://aclanthology.org/2022.findings-emnlp.217}
  {{Conversation Disentanglement with Bi-Level Contrastive Learning}}.
\newblock In \emph{{Findings of the Association for Computational Linguistics:
  EMNLP 2022}}, pages 2985--2996, Abu Dhabi, United Arab Emirates. Association
  for Computational Linguistics.

\bibitem[{Jiang et~al.(2018)Jiang, Chen, Chen, and Wang}]{Jiang2018-gp}
Jyun-Yu Jiang, Francine Chen, Yan-Ying Chen, and Wei Wang. 2018.
\newblock \href {https://aclanthology.org/N18-1164} {{Learning to Disentangle
  Interleaved Conversational Threads with a Siamese Hierarchical Network and
  Similarity Ranking}}.
\newblock In \emph{{Proceedings of the 2018 Conference of the North {A}merican
  Chapter of the Association for Computational Linguistics: Human Language
  Technologies, Volume 1 (Long Papers)}}, pages 1812--1822, New Orleans,
  Louisiana. Association for Computational Linguistics.

\bibitem[{Khadilkar et~al.(2022)Khadilkar, KhudaBukhsh, and
  Mitchell}]{Khadilkar2022-pl}
Kunal Khadilkar, Ashiqur~R KhudaBukhsh, and Tom~M Mitchell. 2022.
\newblock \href
  {https://www.sciencedirect.com/science/article/pii/S266638992100283X}
  {{Gender bias, social bias, and representation: 70 years of Bollywood and
  Hollywood}}.
\newblock \emph{Patterns (New York, N.Y.)}, 3(2):100409.

\bibitem[{Kozloff(2000)}]{Kozloff2000-ox}
Sarah Kozloff. 2000.
\newblock \href {https://play.google.com/store/books/details?id=daUwDwAAQBAJ}
  {\emph{{Overhearing Film Dialogue}}}.
\newblock University of California Press.

\bibitem[{Kummerfeld et~al.(2019)Kummerfeld, Gouravajhala, Peper, Athreya,
  Gunasekara, Ganhotra, Patel, Polymenakos, and
  Lasecki}]{kummerfeld-etal-2019-large}
Jonathan~K. Kummerfeld, Sai~R. Gouravajhala, Joseph~J. Peper, Vignesh Athreya,
  Chulaka Gunasekara, Jatin Ganhotra, Siva~Sankalp Patel, Lazaros~C
  Polymenakos, and Walter Lasecki. 2019.
\newblock \href {https://doi.org/10.18653/v1/P19-1374} {A large-scale corpus
  for conversation disentanglement}.
\newblock In \emph{Proceedings of the 57th Annual Meeting of the Association
  for Computational Linguistics}, pages 3846--3856, Florence, Italy.
  Association for Computational Linguistics.

\bibitem[{Lacan(2006)}]{Lacan2006-xg}
Jacques Lacan. 2006.
\newblock \href {https://play.google.com/store/books/details?id=3LLPB94zclMC}
  {\emph{{Ecrits: The First Complete Edition In English}}}.
\newblock W. W. Norton \& Company.

\bibitem[{Lakoff and Tannen(1984)}]{Lakoff1984-og}
Robin~Tolmach Lakoff and Deborah Tannen. 1984.
\newblock {Conversational strategy and metastrategy in a pragmatic theory: The
  example of Scenes from a Marriage}.
\newblock \emph{Semiotica}, 49(3-4):323--346.

\bibitem[{Lauzen(2019)}]{Lauzen2019-ta}
Martha~M Lauzen. 2019.
\newblock \href
  {http://womenintvfilm.sdsu.edu/wp-content/uploads/2020/01/2019_Its_a_Mans_Celluloid_World_Report_REV.pdf}
  {{It'sa man's (celluloid) world: Portrayals of female characters in the top
  grossing films of 2018}}.
\newblock \emph{Center for the Study of Women in Television and Film, San Diego
  State University}, pages 411--428.

\bibitem[{Lawrence(1991)}]{Lawrence1991-lk}
Amy Lawrence. 1991.
\newblock \href {https://play.google.com/store/books/details?id=6idnyfdp3VIC}
  {\emph{{Echo and Narcissus: Women's Voices in Classical Hollywood Cinema}}}.
\newblock University of California Press.

\bibitem[{Liu et~al.(2020)Liu, Shi, Gu, Liu, Wei, and Zhu}]{Liu2020-rr}
Hui Liu, Zhan Shi, Jia-Chen Gu, Quan Liu, Si~Wei, and Xiaodan Zhu. 2020.
\newblock \href {https://www.ijcai.org/proceedings/2020/535} {{End-to-End
  Transition-Based Online Dialogue Disentanglement}}.
\newblock In \emph{{Proceedings of the Twenty-Ninth International Joint
  Conference on Artificial Intelligence}}, pages 3868--3874, Yokohama, Japan.
  International Joint Conferences on Artificial Intelligence Organization.

\bibitem[{Liu et~al.(2021)Liu, Zhang, Zhao, Zhou, and Zhou}]{Liu2021-uz}
Longxiang Liu, Zhuosheng Zhang, Hai Zhao, Xi~Zhou, and Xiang Zhou. 2021.
\newblock \href {https://ojs.aaai.org/index.php/AAAI/article/view/17582}
  {{Filling the Gap of Utterance-aware and Speaker-aware Representation for
  Multi-turn Dialogue}}.
\newblock \emph{Proceedings of the AAAI Conference on Artificial Intelligence},
  35(15):13406--13414.

\bibitem[{Ma et~al.(2022)Ma, Zhang, and Zhao}]{ma-etal-2022-structural}
Xinbei Ma, Zhuosheng Zhang, and Hai Zhao. 2022.
\newblock \href {https://doi.org/10.18653/v1/2022.acl-long.23} {Structural
  characterization for dialogue disentanglement}.
\newblock In \emph{Proceedings of the 60th Annual Meeting of the Association
  for Computational Linguistics (Volume 1: Long Papers)}, pages 285--297,
  Dublin, Ireland. Association for Computational Linguistics.

\bibitem[{Mahajan and Shaikh(2021)}]{Mahajan2021-ov}
Khyati Mahajan and Samira Shaikh. 2021.
\newblock \href {https://aclanthology.org/2021.sigdial-1.36} {{On the Need for
  Thoughtful Data Collection for Multi-Party Dialogue: A Survey of Available
  Corpora and Collection Methods}}.
\newblock In \emph{{Proceedings of the 22nd Annual Meeting of the Special
  Interest Group on Discourse and Dialogue}}, pages 338--352, Singapore and
  Online. Association for Computational Linguistics.

\bibitem[{Mayfield et~al.(2012)Mayfield, Adamson, and
  Penstein~Ros{\'e}}]{Mayfield2012-us}
Elijah Mayfield, David Adamson, and Carolyn Penstein~Ros{\'e}. 2012.
\newblock \href {https://aclanthology.org/W12-1607} {{Hierarchical Conversation
  Structure Prediction in Multi-Party Chat}}.
\newblock In \emph{{Proceedings of the 13th Annual Meeting of the Special
  Interest Group on Discourse and Dialogue}}, pages 60--69, Seoul, South Korea.
  Association for Computational Linguistics.

\bibitem[{McDaniel et~al.(1996)McDaniel, Olson, and Magee}]{McDaniel1996-gh}
Susan~E McDaniel, Gary~M Olson, and Joseph~C Magee. 1996.
\newblock \href {https://doi.org/10.1145/240080.240187} {{Identifying and
  analyzing multiple threads in computer-mediated and face-to-face
  conversations}}.
\newblock In \emph{{Proceedings of the 1996 ACM conference on Computer
  supported cooperative work}}, CSCW '96, pages 39--47, New York, NY, USA.
  Association for Computing Machinery.

\bibitem[{McKee(2016)}]{McKee2016-fo}
Robert McKee. 2016.
\newblock \emph{{Dialogue: The art of verbal action for page, stage, and
  screen}}.
\newblock Hachette UK.

\bibitem[{Meil{\u a}(2007)}]{Meila2007-pc}
Marina Meil{\u a}. 2007.
\newblock \href
  {https://www.sciencedirect.com/science/article/pii/S0047259X06002016}
  {{Comparing clusterings---an information based distance}}.
\newblock \emph{Journal of multivariate analysis}, 98(5):873--895.

\bibitem[{Nelmes(2010)}]{Nelmes2010-oi}
Jill Nelmes, editor. 2010.
\newblock \href
  {https://www.taylorfrancis.com/books/edit/10.4324/9780203843383/analysing-screenplay-jill-nelmes}
  {\emph{{Analysing the Screenplay}}}.
\newblock Routledge, London, England.

\bibitem[{Ng and Bradac(1993)}]{Ng1993-ju}
Sik~H Ng and James~J Bradac. 1993.
\newblock \href {https://play.google.com/store/books/details?id=xjFiAAAAMAAJ}
  {\emph{{Power in Language: Verbal Communication and Social Influence}}}.
\newblock SAGE Publications.

\bibitem[{O'Brien(2014)}]{OBrien2014-qv}
Anne O'Brien. 2014.
\newblock \href {https://journals.sagepub.com/doi/abs/10.1177/0163443714544868}
  {{`Men own television': why women leave media work}}.
\newblock \emph{Media Culture \& Society}, 36(8):1207--1218.

\bibitem[{O'Meara(2016)}]{OMeara2016-ek}
Jennifer O'Meara. 2016.
\newblock \href {https://doi.org/10.1080/14680777.2016.1234239} {{What ``The
  Bechdel Test'' doesn't tell us: examining women's verbal and vocal
  (dis)empowerment in cinema}}.
\newblock \emph{Feminist Media Studies}, 16(6):1120--1123.

\bibitem[{O'Meara(2019)}]{OMeara2019-jh}
Jennifer O'Meara. 2019.
\newblock \href {https://play.google.com/store/books/details?id=ViG1wgEACAAJ}
  {\emph{{Engaging Dialogue: Cinematic Verbalism in American Independent
  Cinema}}}.
\newblock Edinburgh University Press.

\bibitem[{O'Neill and Martin(2003)}]{ONeill2003-bf}
Jacki O'Neill and David Martin. 2003.
\newblock \href {https://doi.org/10.1145/958160.958167} {{Text chat in
  action}}.
\newblock In \emph{{Proceedings of the 2003 international ACM SIGGROUP
  conference on Supporting group work}}, GROUP '03, pages 40--49, New York, NY,
  USA. Association for Computing Machinery.

\bibitem[{Ramakrishna et~al.(2017)Ramakrishna, Mart{\'\i}nez, Malandrakis,
  Singla, and Narayanan}]{ramakrishna-etal-2017-linguistic}
Anil Ramakrishna, Victor~R. Mart{\'\i}nez, Nikolaos Malandrakis, Karan Singla,
  and Shrikanth Narayanan. 2017.
\newblock \href {https://doi.org/10.18653/v1/P17-1153} {Linguistic analysis of
  differences in portrayal of movie characters}.
\newblock In \emph{Proceedings of the 55th Annual Meeting of the Association
  for Computational Linguistics (Volume 1: Long Papers)}, pages 1669--1678,
  Vancouver, Canada. Association for Computational Linguistics.

\bibitem[{Ravid(1999)}]{Ravid1999-ji}
S~Abraham Ravid. 1999.
\newblock \href {http://www.jstor.org/stable/10.1086/209624} {{Information,
  Blockbusters, and Stars: A Study of the Film Industry}}.
\newblock \emph{The Journal of Business}, 72(4):463--492.

\bibitem[{Richardson(2010)}]{Richardson2010-zp}
Kay Richardson. 2010.
\newblock \href {https://play.google.com/store/books/details?id=onQXUlJ85voC}
  {\emph{{Television Dramatic Dialogue: A Sociolinguistic Study}}}.
\newblock Oxford University Press.

\bibitem[{Roberts(1996)}]{Roberts1996-tc}
Craige Roberts. 1996.
\newblock \href
  {https://www-formal.stanford.edu/buvac/95-context-symposium/Papers/croberts.ps}
  {{Information structure in discourse: Toward a unified theory of formal
  pragmatics}}.
\newblock \emph{Ohio State University Working Papers in Linguistics},
  49:91--136.

\bibitem[{Rosen(1973)}]{rosen73}
Marjorie Rosen. 1973.
\newblock \emph{Popcorn Venus; Women, Movies and the American Dream}.
\newblock Coward, McCann and Geoghegan.

\bibitem[{Sacks et~al.(1974)Sacks, Schegloff, and Jefferson}]{Sacks1974-cg}
Harvey Sacks, Emanuel~A Schegloff, and Gail Jefferson. 1974.
\newblock \href {http://www.jstor.org/stable/412243} {{A Simplest Systematics
  for the Organization of Turn-Taking for Conversation}}.
\newblock \emph{Language}, 50(4):696--735.

\bibitem[{Sang et~al.(2022)Sang, Mou, Yu, Yao, Li, and Stanton}]{Sang2022-ao}
Yisi Sang, Xiangyang Mou, Mo~Yu, Shunyu Yao, Jing Li, and Jeffrey Stanton.
  2022.
\newblock \href {https://aclanthology.org/2022.naacl-main.317} {{TVShowGuess:
  Character Comprehension in Stories as Speaker Guessing}}.
\newblock In \emph{{Proceedings of the 2022 Conference of the North American
  Chapter of the Association for Computational Linguistics: Human Language
  Technologies}}, pages 4267--4287, Seattle, United States. Association for
  Computational Linguistics.

\bibitem[{Sap et~al.(2017)Sap, Prasettio, Holtzman, Rashkin, and
  Choi}]{sap-etal-2017-connotation}
Maarten Sap, Marcella~Cindy Prasettio, Ari Holtzman, Hannah Rashkin, and Yejin
  Choi. 2017.
\newblock \href {https://doi.org/10.18653/v1/D17-1247} {Connotation frames of
  power and agency in modern films}.
\newblock In \emph{Proceedings of the 2017 Conference on Empirical Methods in
  Natural Language Processing}, pages 2329--2334, Copenhagen, Denmark.
  Association for Computational Linguistics.

\bibitem[{Shen et~al.(2006)Shen, Yang, Sun, and Chen}]{Shen2006-qc}
Dou Shen, Qiang Yang, Jian-Tao Sun, and Zheng Chen. 2006.
\newblock \href {https://doi.org/10.1145/1148170.1148180} {{Thread detection in
  dynamic text message streams}}.
\newblock In \emph{{Proceedings of the 29th annual international ACM SIGIR
  conference on Research and development in information retrieval}}, SIGIR '06,
  pages 35--42, New York, NY, USA. Association for Computing Machinery.

\bibitem[{Silverman(1988)}]{Silverman1988-dl}
Kaja Silverman. 1988.
\newblock \emph{{The Acoustic Mirror: The Female Voice in Psychoanalysis and
  Cinema}}.
\newblock Indiana University Press.

\bibitem[{Sjogren(2006)}]{Sjogren2006-rj}
Britta~H Sjogren. 2006.
\newblock \href {https://openlibrary.org/books/OL18208503M.opds} {\emph{{Into
  the vortex female voice and paradox in film}}}.
\newblock University of Illinois Press, Urbana.

\bibitem[{Wang et~al.(2020)Wang, Hoi, and Joty}]{Wang2020-pz}
Weishi Wang, Steven C~H Hoi, and Shafiq Joty. 2020.
\newblock \href {https://aclanthology.org/2020.emnlp-main.533} {{Response
  Selection for Multi-Party Conversations with Dynamic Topic Tracking}}.
\newblock In \emph{{Proceedings of the 2020 Conference on Empirical Methods in
  Natural Language Processing (EMNLP)}}, pages 6581--6591, Online. Association
  for Computational Linguistics.

\bibitem[{Wu et~al.(2020)Wu, Hoi, Socher, and Xiong}]{Wu2020-oq}
Chien-Sheng Wu, Steven C~H Hoi, Richard Socher, and Caiming Xiong. 2020.
\newblock \href {https://aclanthology.org/2020.emnlp-main.66} {{{TOD}-{BERT}:
  Pre-trained Natural Language Understanding for Task-Oriented Dialogue}}.
\newblock In \emph{{Proceedings of the 2020 Conference on Empirical Methods in
  Natural Language Processing (EMNLP)}}, pages 917--929, Online. Association
  for Computational Linguistics.

\bibitem[{Xu et~al.(2021)Xu, Jie, Lu, and Bing}]{xu-etal-2021-better}
Lu~Xu, Zhanming Jie, Wei Lu, and Lidong Bing. 2021.
\newblock \href {https://doi.org/10.18653/v1/2021.naacl-main.271} {Better
  feature integration for named entity recognition}.
\newblock In \emph{Proceedings of the 2021 Conference of the North American
  Chapter of the Association for Computational Linguistics: Human Language
  Technologies}, pages 3457--3469, Online. Association for Computational
  Linguistics.

\bibitem[{Yu and Joty(2020)}]{yu-joty-2020-online}
Tao Yu and Shafiq Joty. 2020.
\newblock \href {https://doi.org/10.18653/v1/2020.emnlp-main.512} {Online
  conversation disentanglement with pointer networks}.
\newblock In \emph{Proceedings of the 2020 Conference on Empirical Methods in
  Natural Language Processing (EMNLP)}, pages 6321--6330, Online. Association
  for Computational Linguistics.

\bibitem[{Zhu et~al.(2020)Zhu, Nan, Wang, Nallapati, and Xiang}]{Zhu2020-hx}
Henghui Zhu, Feng Nan, Zhiguo Wang, Ramesh Nallapati, and Bing Xiang. 2020.
\newblock \href {https://ojs.aaai.org/index.php/AAAI/article/view/6524} {{Who
  Did They Respond to? Conversation Structure Modeling Using Masked
  Hierarchical Transformer}}.
\newblock \emph{Proceedings of the ... AAAI Conference on Artificial
  Intelligence. AAAI Conference on Artificial Intelligence}, 34(05):9741--9748.

\bibitem[{Zhu et~al.(2021)Zhu, Lau, and Qi}]{Zhu2021-yq}
Rongxin Zhu, Jey~Han Lau, and Jianzhong Qi. 2021.
\newblock \href {https://aclanthology.org/2021.alta-1.1} {{Findings on
  Conversation Disentanglement}}.
\newblock In \emph{{Proceedings of the The 19th Annual Workshop of the
  Australasian Language Technology Association}}, pages 1--11, Online.
  Australasian Language Technology Association.

\end{thebibliography}
\bibliographystyle{acl_natbib}

\clearpage
\newpage
\appendix

\section{Evaluation metrics}
\label{sec:metrics}

\begin{itemize}
    \item{\textbf{Adjusted Random Index (ARI)}~\cite{Halkidi2002-rl} is defined as:
    
 {\small\begin{equation}
    \frac{\sum_{i,j} \binom{n_{i,j}}{2} - [\sum_i \binom{a_i}{2} \sum_j \binom{b_j}{2}] / \binom{n}{2}}{\frac{1}{2} [\sum_i \binom{a_i}{2} + \sum_j \binom{b_j}{2}] - [\sum_i \binom{a_i}{2} \sum_j \binom{b_j}{2}] / \binom{n}{2}}
   \end{equation}\normalsize}}
   
    \item \textbf{Variation of Information (VI)} is the information gain or loss when going from one clustering to another \cite{Meila2007-pc}. It is the sum of conditional entropies $H(Y |X) + H(X|Y )$, where $X$ and $Y$ are clusters of the same set of items. We report $1 - $VI, so the larger the value the better.
    \item{\textbf{Shen F\textsubscript{1}}~\cite{Shen2006-qc}: 
    Given a detected thread $j$ and true thread $i$: 
    
    {\small\begin{align}
        F(i, j) &= \frac{2\times \text{Precision}(i, j) \times \text{Recall}(i, j)}{\text{Precision}(i, j)+\text{Recall}(i, j)} \\
        F &= \sum_i \frac{n_i}{n} \max_{j} F(i, j)
    \end{align}\normalsize}
    
    where Recall$(i, j) = \tfrac{n_{i,j}}{n_i}$, Precision$(i, j) = \tfrac{n_{i,j}}{n_j}$, $n_{i,j}$ is the number of messages of thread $i$ in $j$, $n_j$ the number of messages in detected thread $j$, $n_i$ thread $i$.}
    \item \textbf{One-to-One Overlap}~\cite{elsner-charniak-2008-talking} calculates the percentage of overlap between two sets of conversational threads paired up with the max-flow algorithm.
   \item \textbf{Exact Match F\textsubscript{1}}~\cite{kummerfeld-etal-2019-large} calculates the number of perfectly matched conversational threads between two sets. During our annotation process, we did see threads with only one dialogue line, which functions quite differently from system messages in the context of IRC, so we did not exclude conversations with only one dialogue line.
    
\end{itemize}

\section{Annotation guidelines}
\label{sec:annotation-guidelines}

\subsection{Building intuitions}

This is a study of conversational behaviors of characters in drama: here, we consider TV shows and movies and the specific task of conversation disentanglement. 
On the highest level, we want to identify \textit{threads} in a conversation between multiple characters in a scene of a TV show or movie. 
In the same scene, some characters can change the subject of a conversation, or redirect other characters' attention to themselves, while others might never do so. 
Characters in a closer relationship might converse more frequently with each other. 
We annotate to build a dataset that can help us investigate those inquiries. To (hopefully) ease understanding, all examples below are drawn from \textit{Gilmore Girls}, which follows the story of Lorelai Gilmore and her daughter, Rory, in a small town in Connecticut. In the excerpts below, we see Luke, Lorelai's will-they-or-won't-they love interest throughout the series, and Emily, Lorelai's mother. \textit{Gilmore Girls} is famous for its fast-paced dialogue and offers illustrations of conversational behaviors that we wish to study.

\subsubsection{Dialogue, conversation, and reply-to}

\paragraph{Dialogue as speech act.} Our definition of dialogue is an all-encompassing one taken from \citet[p. 2]{McKee2016-fo}: ``Any words said by any character to anyone.'' A line of dialogue is, then, a sequence of words uttered (or, an \textit{utterance}) by a character to themselves, another character, or a few other characters. This view, unlike a narrower one, where dialogue is a conversation held between characters, sees dialogue as a verbal tactic initialized by a character to achieve a certain goal: ``All talk responds to a need, engages a purpose, and performs an action''~\citet[p. 3]{McKee2016-fo}. In other words, a dialogue is a \textit{speech act}. Characters use dialogue to inform us of an event (exposition), tell us something about themselves (characterization), or try to make something happen (action). 

\paragraph{Conversation and conversational thread.} We adopt a broad definition of a conversation: conversation is a talk between characters. In defining thread, we first adapt \citeposs{Goffman1963-ze}'s definition of \textit{conversation}\footnote{We note that thread in other work might be called \textit{sub-conversation}, or simply \textit{conversation}, as opposed to a stream of messages described in e.g.,~\citealp{elsner-charniak-2008-talking}. Given the complexity of drama, we chose not to overload the term of \textit{conversation} and preferred \textit{thread}, which also captures their interleaving nature.}: a thread is a kind of \textit{focused interaction}: one where ``persons gather close together and openly cooperate to sustain a single focus of attention, typically by taking turns at talking''~\cite[p. 24]{elsner-charniak-2008-talking}. Second, we assume conversations in a given scene in drama are \textit{entangled} and there are often more than one \textit{thread} in any given conversation. Taken together, here is our operative definition:

\begin{quote}
A thread is a cluster of semantically and pragmatically coherent utterances that are part of a conversation. Those utterances share a single, sustainable focus of attention~\cite{Goffman1963-ze,Ervin-Tripp1964-dq,Sacks1974-cg}, either on a character (who has other characters' attention) or a topic (often related to the wants and needs of a character), as well as other observable contextual relations~\cite{ONeill2003-bf}.
\end{quote}

\noindent In a conversation, attention can be paid to a character (who has the floor) or a topic (why they are having this conversation). In the context of drama, we can often relate the topic of a thread to \textbf{the desire or intent of the character who started the conversation} (this loops back to McKee's idea that dialogues are speech acts): characters can express their own needs, or respond to someone else's needs, and the need acts as the driving passion of the dramatic conversation (and subsequently, plot, characterization, etc.).  

Attention can be paid to a character or a topic, but such attention should be \textit{sustainable} over multiple utterances to form a thread. In other words, when an utterance redistributes the focus of other characters (or, a change of \textit{floor}) or shows us new wants and needs of a character, and such distribution or attention or topical focus is carried over to the next couple of utterances, it usually marks the start of a new conversational thread. However, there are occasions where threads can be short, which we will describe at the end of this section.

It has been noted that the exact start of a conversational thread is not easy to determine~\cite{McDaniel1996-gh,Elsner2010-wi}, so we will dedicate the second section (\S A.1.2) to this topic. There, we try to unpack our operational definition with more examples. 

\paragraph{Reply-to relationship.} Following conventions in NLP~\cite{Zhu2021-yq}, we understand individual dialogues in a multi-party conversation in terms of \textit{parent utterance} and \textit{utterance of interest} (UOI), and \textit{utterance} is roughly synonymous with a dialogue line. The following is the first two utterances in the entire series of \textit{Gilmore Girls}:

\begin{quote}
    \begin{itemize}
        \item[{\scshape lorelai}] Please, Luke. Please, please, please.
        \item[{\scshape luke}] How many cups have you had this morning?
        \item[] {\hfill(I.i)\footnote{Season 1, episode 1 from \textit{Gilmore Girls.} We are adapting the olden MLA convention for Shakespearean plays, where uppercase Roman numerals denote season, and lowercase ones, episode.}}
    \end{itemize}
\end{quote}

\noindent In practice, UOI means the line you are currently annotating, and its parent utterance is the previous line it most logically \textit{replies to}. In the example above, if Luke's utterance is the UOI, its parent utterance is the immediately previous utterance by Lorelai (``Please, Luke \dots''). In fact, the \textit{default} parent utterance is usually the previous line. Perhaps we can think of conversations in drama as sequences of sentences, where one \textit{triggers} the next. Given our qualitative observations of drama, we make the following remarks on UOI, parent utterance, and thread:

\begin{itemize}
    \item If an UOI does not have any parent utterance, it's the start of a new thread.
    \item One UOI can only have ONE parent utterance. An utterance can have multiple \textit{children} (utterances that point to it as parent utterance).
    \item The default parent utterance is the previous utterance.
\end{itemize}

\paragraph{Nuances of reply-to and sentences in one dialogue turn.}  Those preliminary remarks don't always apply. Often one dialogue line is too large a unit for us to fully understand conversational behaviors, and more often the next line isn't the response to the current: %

\begin{quote}
    \begin{itemize}
\item[{\scshape emily}] You're being stubborn, as usual.
\item[{\scshape lorelai}] No, Mom, I'm not being stubborn. I'm being me! The same person who always needed to work out her own problems and take care of herself. Because that's the way I was born. That's how I am!
\item[{\scshape emily}] Florence, I'm dripping.
\item[{\scshape lorelai}] I appreciate what you have done for Rory in paying for this school. That will not be forgotten. You won't let it. But she is my daughter. And I decide how we live, not you. Now then, do they validate parking here?
    \item[] {\hfill(I.ii)}
    \end{itemize}
\end{quote}

\noindent Here, ``Florence, I'm dripping.'' is certainly not a reply to the previous line. And even within one dialogue turn, ``Now then, do they validate parking here?'' has nothing to do with whether Lorelai is stubborn or not (which we tentatively call a \textit{topic}). To reflect this, in this work, we study utterances on the \textit{sentence} level. In other words, a screenplay/teleplay exhibits the following hierarchy:

\begin{quote}
    title $>$ scene $>$ conversational thread $>$ dialogue turn $>$ dialogue line/utterance $>$ sentence
\end{quote}

\paragraph{Continuation as reply-to.} Lorelai's long speech has multiple sentences. Given our design choice, we will say that the default parent utterance is still mostly \textit{previous} sentence. The notable exception is, of course, ``Now then, do they validate parking here?''---it \textit{does not} have a parent utterance; it's starting a new thread. To distinguish between a true reply and continuation, we will encode speaker and turn information later in the model. Qualitatively, though, it is reasonable to treat continuation as a special case of reply-to. Consider:

\begin{quote}
    \begin{itemize}
        \item[{\scshape sookie}] And if we go down after two years~\dots
        \item[{\scshape lorelai}] It'll be the most exciting two years of our lives!
        \item[] {\hfill(II.viii)}
    \end{itemize}
\end{quote}

\noindent In conversations, one person can \textit{finish} each other's lines. So, we say, for multiple sentences in one dialogue turn, the parent utterance of sentence $n$ is, by default, sentence $n-1$. 

\paragraph{Look for parent utterances, not addressee.} In determining the reply-to relationship between utterances, the speaker/addressee information can help you find the parent utterance of an UOI, but it shouldn't be your sole basis of judgment, because they are fundamentally different task: determine the relationship between utterances vs. the person to whom a speaker addresses. Consider the following example: 

\begin{quote}
    \begin{itemize}
        \item[{\scshape lorelai}] Michel, come on, we've got to get into these budgets.
        \item[{\scshape sookie}] Now.
        \item[{\scshape michel}] Does the red light mean it's programmed?
        \item[{\scshape sookie}] {[\textit{to Lorelai}]} I explained it a hundred times.
        \item[{\scshape lorelai}] Michel, you've been setting that machine for 20 minutes now.
        \item[] {\hfill(IV.xvi)}
    \end{itemize}
\end{quote}

\noindent According to the action statement, Sookie says ``I explained it a hundred times.'' to Lorelai. But that line is \textit{triggered} by Michel's ``Does the red light mean it's programmed?'' So, we say Sookie's ``I explained it a hundred times'' is the parent utterance of Michel's ``Does the red light mean it's programmed?''---even though she says that directly to Lorelai. If this seems odd, remember we cannot assume Michel does not hear Sookie's line, and that if we were to say Sookie replied to Lorelai, there's no line from Lorelai that can reasonably be the parent utterance.

\paragraph{Intuition: thread, topic, and floor} Before we define a thread more thoroughly, we use two examples to drive the intuition. If the next section (\S A.1.2) gets confusing, revisit those two examples.

A \textit{topic} is a semantically and pragmatically coherent unit:

\begin{quote}
    \begin{itemize}
        \item[{\scshape lorelai}] Who is that?
        \item[{\scshape rory}] I don't know. She just followed me in here like a puppy dog without saying a word.
        \item[{\scshape lorelai}]  Maybe she's lost.
        \item[{\scshape rory}] Or, maybe she's one of my new suitemates who I'm already off to a swell start with. 
        \item[{\scshape lorelai}]  Do you know how vulnerable you are to venereal disease?
        \item[{\scshape rory}]  All hail to the queen of the nonsequiturs.
        \item[{\scshape lorelai}] This parent orientation I went to was a nonstop litany of the horrors awaiting college freshman. You're supposed to carry a whistle, a flashlight, a crucifix, and a loaded Glock with you at all times.
        \item[{\scshape rory}] We should go out there. She'll think we're hiding.
        \item[{\scshape lorelai}] Okay, just don't shake hands with her. Bacteria.
        \item[{\scshape rory}] Mom.
        \item[{\scshape lorelai}] Or tell her where you live.
        \item[{\scshape rory}] Too late.
        \item[{\scshape lorelai}] Oh, you touched the doorknob.
        \item[{\scshape rory}] Good grief.
        \item[] {\hfill(IV.ii)}
    \end{itemize}
\end{quote}

\noindent We first note that topic is not entirely about coherence from the semantic point of view: we expect to see metaphors or jokes in TV shows and movies, and there's no puppy dog, queen, or crucifix really present in the scene; and just like in real life, we might bring up something that appears entirely random in any conversation (``You have to be in my brain to see the connection!'') Still, we can intuitively tell there are two threads of conversation going on, and we aren't perplexed when Rory says ``All hail to the queen of nonsequiturs.'' We can perhaps look at this previous exchange from a pragmatic perspective: Lorelai starts that ``Who is this'' conversation because she wants to find out who this girl in Rory's dorm room is. Venereal disease has nothing to do with that. It \textit{does not follow}. 

On the other hand, \textit{floor} has to do with \textit{ownership} and \textit{attention}. Who has started and owns the conversation? Who controls our (and other characters') attention? Who do we direct our gaze to if we were also in the scene? 

\begin{quote}
    \begin{itemize}
        \item[{\scshape lorelai}] You could tape the movies, or get a DVD player.
        \item[{\scshape emily}] I don't need a DVD player. 
        \item[{\scshape lorelai}] Well, why not? Then you could buy all those musicals you love and watch them whenever you felt like it. 
        \item[{\scshape emily}] I'm not an invalid, Lorelai. 
        \item[{\scshape lorelai}] Well, of course you are, Mother. Why else would I suggest a DVD player? 
        \item[{\scshape emily}] I can fill my time all by myself and I'd like you to drop this conversation right now.
        \item[{\scshape lorelai}] Where are you going?
        \item[{\scshape emily}] We're going to eat.
        \item[{\scshape lorelai}] Just because you leave the room doesn't mean the conversation's over. I started the conversation. The conversation's in me. Therefore, when I get over there, the conversation's just gonna start up again!
        \item[] {\hfill(III.xiii)}
    \end{itemize}
\end{quote}

\noindent Lorelai's last line can help us understand the nature of a conversational thread: a specific character first \textit{started} and \textit{owned} the thread, and then they let someone else do so. One character is ready to speak or stays speaking, and others listen and reply. Here, Lorelai wants to buy Emily a DVD player, and the latter refuses. Emily leaving the room can be understood as her unwilling to pay more attention to Lorelai: Emily doesn't want to stay in the conversation anymore. Also, Lorelai asking ``Where are you going'' is a shift in topic: it has nothing to do with DVD players.

We will emphasize this again, but floor change tends to be more common when the thread involves at least three characters. Floor still exists between two-party conversations: if a high school student finds herself in the principal's office, we can expect the principal might be the one doing most of the talking and \textit{has the floor}. But floor change should be more frequent when we have more than two characters.

This exchange between Emily and Lorelai is also an interesting instance where Emily tries to gain control over the conversation (or floor) and Lorelai doesn't let her, so she just leaves. This behavior of floor gaining and changing is a particularly interesting aspect we want to consider for analysis. In the world of \textit{Gilmore Girls}, Emily is the matriarch of the family, and she does have the most \textit{power} in her conversations with Lorelai (her daughter) and Rory (her granddaughter). Who tends to gain the floor? How often does a new thread start? How long is a thread? Those are interesting empirical questions we hope this work can eventually help us answer.

Let's try another one and revisit this example:

\begin{quote}
    \begin{itemize}
        \item[{\scshape lorelai}] (D1) Michel, come on, we've got to get into these budgets.
        \item[{\scshape sookie}] (D2) Now.
        \item[{\scshape michel}] (D3) Does the red light mean it's programmed?
        \item[{\scshape sookie}] (D4) {[\textit{to Lorelai}]} I explained it a hundred times.
        \item[{\scshape lorelai}] (D5) Michel, you've been setting that machine for 20 minutes now.
        \item[] {\hfill(IV.xvi)}
    \end{itemize}
\end{quote}

\noindent Here, we have two threads: thread one has D1, D2, D5 (topic: budget meeting); thread two has D3 and D4 (topic: Michel's recording device). D5 relates most strongly to Lorelai's desire to get Michel to join the meeting, which is expressed already in D1, so it takes precedence over the weak semantic relation of \textit{it} (D3) and \textit{that machine} (D5) and makes D5 part of the first thread.

Here's one last example for intuition. This is a family dinner scene. Rory, the granddaughter, brings her boyfriend, Dean, along, who meets Richard, the grandfather, for the first time. Interrogation ensues. Pay attention to the intention and desire of each character, and who has the control of where the conversation is going.

\begin{quote}
    \begin{itemize}
\item[{\scshape emily}] Antonia, please bring out the Twinkies. 
\item[{\scshape lorelai}] I can't believe I just heard you say those words. 
\item[{\scshape emily}] Well, don't get used to it. 
\item[{\scshape richard}] So, Dean, where are you planning to go to college? 
\item[{\scshape dean}] Oh, uh, well I. . . 
\item[{\scshape lorelai}] Geez Dad, start off with ``what's your favorite baseball team'' or something. 
\item[{\scshape richard}] I'm talking to Dean. 
\item[{\scshape dean}] I don't know yet. 
\item[{\scshape richard}] You don't? 
\item[{\scshape dean}] No, not yet. 
\item[{\scshape richard}] Well, what kind of grades do you get? 
\item[{\scshape emily}] Richard please, don't grill the boy. 
\item[{\scshape richard}] I'm not grilling the boy Emily. It's an easy question. A's, B's, C's? 
\item[{\scshape dean}] I get a mixture actually. 
\item[{\scshape richard}] Mixture? [\textit{laughs}] What's the ratio? 
\item[{\scshape emily}] Richard. 
\item[{\scshape richard}] I'm just trying to get to know the boy Emily. After all, Rory brings home a young man to dinner, the least we can do is learn something about him. 
\item[{\scshape lorelai}] He changes a mean water bottle. 
\item[{\scshape dean}] I get a couple A's, couple B's, few C's. 
\item[{\scshape richard}] Really? 
\item[{\scshape dean}] I'm not great in math. 
\item[{\scshape lorelai}] Yeah, except who is really? You know, except mathematicians or the blackjack dealers, or I guess Stephen Hawking doesn't suck, but you know. You know what else is good though Mom, is a Ho-Ho. Because if you can't find a Twinkie, you know, treat yourself to a nice Ho-Ho. How long does it take to open a box? 
\item[{\scshape emily}] She's making them. 
\item[{\scshape lorelai}] She's making the Twinkies? You're kidding. 
\item[{\scshape emily}] Oh Richard, wasn't there a book you wanted to give Rory? 
\item[{\scshape richard}] In a minute. So Dean \dots
\item[{\scshape rory}] Uh, Grandpa? 
\item[{\scshape richard}] You do know that Rory is going to an Ivy League school? 
\item[{\scshape dean}] I know. 
\item[{\scshape richard}] Harvard, Princeton, Yale. 
\item[{\scshape lorelai}] He said he knew, Dad. 
\item[{\scshape richard}] You need top grades to get into a top school. 
\item[{\scshape dean}] Yeah, well, Rory's really smart. 
\item[{\scshape richard}] Yeah, she is really smart. 
\item[{\scshape rory}] Mom? 
\item[{\scshape lorelai}] Yeah, why don't we all go sit in the uh \dots
\item[{\scshape richard}] So, how are you planning to make a living once you graduate from this college you haven't thought anything about yet? 
\item[{\scshape rory}] Grandpa, can we talk about something else? 
\item[{\scshape emily}] I'm going to get that book.
        \item[] {\hfill(II.i)}
    \end{itemize}
\end{quote}

\noindent There are two threads: one about Twinkie, the other about Richard's approval (or the lack thereof) of Dean being his granddaughter's boyfriend. For most of this excerpt, Richard has the floor. Lorelai, Emily, and Rory all try to take over (or, gain the floor) and stop Richard, and all fail. Semantically, you can say that Richard asks questions that span through a few topics---his college plans, his grades at school, Rory's college plans, his career plans (or whether he could provide for Rory). But pragmatically, all those utterances are for Richard to get to know Dean. Each question, then, will not start a new thread.

\paragraph{Threads can be short.} While they are mostly organized by a sustainable distribution of attention, threads don't have to be long. There can be \textit{hanging threads}, where one character tries to switch subject or gain floor but failed. If an utterance is a reply to an action, the thread might be short as well. Here are two examples that illustrate that:

\begin{quote}
    \begin{itemize}
        \item[{\scshape rory}] Was anything resolved? Are she and grandpa gonna be all right?
        \item[{\scshape lorelai}] Don't worry about it. They're a team. They'll be okay.
        \item[{\scshape rory}]  Good. I like them. [\textit{begins eating with fingers}] 
        \item[{\scshape lorelai}] I know. [\textit{takes a bite}] 
        \item[]{[\textit{Luke brings plates and forks and transfers napkins to plates}]}
        \item[{\scshape rory}] Thanks.
        \item[]{[\textit{Lorelai pulls out her rosebud and hands it to Luke. While they resume eating, Luke walks away sniffing the rose.}]}
        \item[] {\hfill(III.xiii)}
    \end{itemize}
\end{quote}

\noindent Rory and Lorelai were talking about Rory's grandparents at first. From the action statements, we see Luke brings plates and forks to the table, and Rory's ``Thanks'' at the end replies to that. It does not belong in the previous thread about her grandparents, and here the only logical annotation would indicate that Rory starts a new thread, which has only one sentence, before the scene ends.

\begin{quote}
    \begin{itemize}
        \item[{\scshape emily}] We were buying the two of you a house. Doesn't the fact that we were willing to spend an enormous amount of money on a wedding present count for anything?
        \item[{\scshape lorelai}] So, that's what you're mad about? Your mad about the enormous amount of money you might have wasted?
        \item[{\scshape rory}] Mom.
        \item[{\scshape emily}] That's not what I was saying. 
        \item[{\scshape lorelai}] Well, you implied it.
        \item[{\scshape emily}] Lorelai, that's---
        \item[{\scshape rory}] Bangalore! Bangalore! Bangalore!
        \item[] {\hfill(VII.iii)}
    \end{itemize}
\end{quote}

\noindent In this scene, Emily and Lorelai are having a quarrel. Rory tries to distract everyone and change topic, yelling ``Bangalore'' entirely out of the blue. Rory's ``Bangalore'' also starts a new thread. 

In both examples, both threads have only one utterance, which might be shorter than usual, but contextually, it makes sense.

In the next section, we will talk in detail about how to determine topic/floor change or thread start in a more principled manner. But ultimately you need to use your judgment, and hopefully you have a rough sense of what a topic or a floor is. 

\subsubsection{Threading drama}

This section offers more instructions on how to decide when a conversational thread starts in drama. We already noted the two layers at work in the previous section: 

\begin{enumerate} 
\item Conceptually, the start of a thread introduces a new floor (``focus of attention'', \citealp{Goffman1963-ze}) or an observable change in topic (contextual relations, \citealp{ONeill2003-bf}). 
\item Practically, the start of a thread has no parent utterance. 
\end{enumerate} 

\noindent The rest of this section further elaborates our (perhaps idiosyncratic) definitions of floor and topic as they apply to our conversational analysis of drama.

\paragraph{Floor: Who are we paying attention to?} 

\citet[p. 392]{Elsner2010-wi} describe the start of a new conversational thread as the process of participants (or in our case, characters) ``hav[ing] refocused their attention \dots away from whoever held the floor in the parent conversation.'' Like in \citet{Goffman1963-ze}, \textit{attention} is the operative word: the character holding the floor can safely assume that they have attention from others. Such attention is singular and must be sustained throughout the thread; or someone else has the floor. A test we could use is whether we could logically insert ``Now, everyone listen to me!'' or ``Now, everyone look at me!'' before the UOI that we see as a potential start of a new thread. 

Consider the following example:

\begin{quote}
    \begin{itemize}
    \item[{\scshape taylor}] This goes well beyond a head of lettuce, young man. The charges against your nephew are numerous. He stole the ``save the bridge'' money!
    \item[{\scshape luke}] He gave that back.
    \item[{\scshape taylor}] He stole a gnome from Babette's garden.
    \item[{\scshape luke}] Pierpont was also returned.
    \item[{\scshape miss patty}] He hooted at one of my dance classes.
    \item[{\scshape fran}] He took a garden hose from my yard.
    \item[{\scshape andrew}] My son said he set off the fire alarms at school last week.
    \item[{\scshape lorelai}] I heard he controls the weather and wrote the screenplay to \textit{Glitter}. 
    \item[{\scshape bootsy}] {\small\sffamily[Now, everyone listen to me:]} I think it's time for me to pipe up here.
    \item[{\scshape luke}] Oh yeah, that'll be good.
    \item[{\scshape bootsy}] I have every right to pipe in here, Luke. I'm a local entrepreneur.
    \item[{\scshape luke}] You took over your father's newsstand, Bootsy. It doesn't make you an entrepreneur.
    \item[{\scshape bootsy}] And you took over your old man's hardware store.
    \item[{\scshape luke}] And turned it into a diner!
    \item[{\scshape bootsy}] Big whoop. Who can't fry an egg?
    \item[{\scshape taylor}] Let's keep it moving here boys, huh?
    \item[{\scshape bootsy}] I never liked the look of that kid from the second I saw him.
    \item[{\scshape luke}] Unbelievable.
    \item[{\scshape bootsy}] Excuse me, but I've got the floor.
    \item[{\scshape luke}] You don't have the floor.
    \item[{\scshape bootsy}] I'm standing, aren't I?
    \item[{\scshape luke}] Well, I was standing first, which means I have the floor and I'm not giving it to you.
    \item[{\scshape taylor}] What is with you two? 
    \item[] {\hfill(II.viii)}
    \end{itemize}
\end{quote}

\noindent It should be clear that by saying ``I think it's time for me to pipe up here,'' Bootsy is trying to get everyone else's attention. And since he is able to keep talking, we see he manages to sustain the attention and (unfortunately, \textit{contra} Luke) hold the floor, which is very different from the utterances of Miss Patty, Fran, and Andrew: they \textit{interject}, but they do not gain the floor. In this light, threads are usually composed of several dialogue turns because that is the only way for us to tell whether such attention lasts (unless, of course, the scene ends abruptly).

\paragraph{Topic: What does the character want?} 

As we've seen in \citealp{McKee2016-fo}: ``All talk responds to a need, engages a purpose, and performs an action.'' With that in mind, when you see the first utterance in the scene, try to identify that need, purpose, or action it speaks to, which you can describe in a \textit{verb} phrase. From there, extrapolate a broader, general topic, which you can describe in a \textit{noun} phrase. Remember, you might not be able to do such extrapolation without knowing what's happening in the scene. So read ahead and skim a little bit. A new conversational thread begins when the UOI has nothing to do with that previous topic. Otherwise, it continues the current thread. %

A simple test to see if this UOI starts a new topic is to insert such parenthetical statements as ``Changing subject'' at the beginning, and see if the conversation still makes sense.  On the flip side, we, adapting \citet{McDaniel1996-gh}, introduce some basic semantic and pragmatic mechanisms that signal the \textit{continuation} of a thread below:

\begin{itemize}
    \item \textit{semantically and pragmatically coherent response}: By definition, if the UOI makes a coherent response to any previous lines, the closest candidate to it is its parent utterance.
    \item \textit{semantically non-topical speech}: Expressive speeches (``Ouch'') or greetings (``Hi'') do not have a ``manifest topic''~\cite{Ervin-Tripp1964-dq}, so we consider them as non-topical. Unless they are used to gain floor, they will always be regarded as a continuation of the same thread.
    \item \textit{successive greetings}: If there are other things going on in the scene, ``Hi'' and ``Goodbye'' \textit{alone} do not form a thread.
    \item \textit{co-reference}: Pronouns whose referent would be less ambiguous if we determine that the UOI continues the current thread.
    \item \textit{term of address}: If one character addresses another directly, there's a higher chance that they are in the same thread.
    \item \textit{acknowledgement}: If one character acknowledges, in their utterance, the presence of another character in the same scene, they are more likely in the same thread.
    \item \textit{same physical location}: If all characters stay where they are at the start of the thread, they might be in the same thread. \textbf{In contrast}: of course, hanging up the phone, someone leaving or entering the room/scene, are useful signals that indicate a new thread might be about to start.
\end{itemize}

\subsubsection{Examples}

We offer three extended examples to motivate threading in drama. Parenthetical statements are in brackets; additional comments are prefixed by the pound (\#) sign:

\paragraph{Example 1. Focus on how Emily keeps changing subject.}

\begin{quote}
    \begin{itemize}
        \item[{\scshape lorelai}] [{\sffamily\small Now, tell me:}] Why are you smiling like that? 
         {\sffamily\small\begin{itemize}
                 \item[\#] Lorelai's purpose: find out why Emily ``smiles like that'' $\rightarrow$ topic: Emily's smile
                 \end{itemize}}
        \item[{\scshape emily}] What are you talking about? 
        \item[{\scshape lorelai}] You're smiling. 
        \item[{\scshape emily}] I'm happy. 
        \item[{\scshape lorelai}] That's not your ``I'm happy'' smile. 
        \item[{\scshape emily}] Well, what smile is it, Loerlai? 
        \item[{\scshape lorelai}] That's your ``I've got something on Lorelai'' smile. 
        \item[{\scshape emily}] [{\sffamily\small Changing subject:}] Rory, your mother must be very tired.
        \item[{\scshape rory}]  She works a lot. 
        \item[{\scshape lorelai}] I grew up with that smile---I know that smile. 
        \item[{\scshape emily}] [{\sffamily\small Changing subject:}] Tell me about school.
        \item[{\scshape rory}]  Well, my French final went pretty well. 
        \item[{\scshape lorelai}] You can change the subject. I know the smile. 
        \item[{\scshape emily}] Whatever you say, dear.\\ {\sffamily\small \# cotinuation: acknowledgement} 
        \item[{\scshape lorelai}] I've used it a few times myself. 
        \item[{\scshape rory}]  Mom. \\
        {\sffamily\small\# continuation: direct address}
        \item[{\scshape emily}] [{\sffamily\small Changing subject:}] So, tell me about parents' day. 
        \item[{\scshape lorelai}] What? 
        \item[{\scshape emily}] Parent's day? Next Wednesday? When all the parents are supposed to go to the classes with their children all day long? 
        \item[{\scshape lorelai}] The Chilton newsletter came out today!\\  %
        {\sffamily\small\begin{itemize}
         \item[\#] Chilton is the name of Rory's school, which might not be immediately obvious without prior knowledge of the show. But if you read the rest of the scene, or the excerpt here, you should be able to see Chilton is related to the ``smile'' and ``school'' threads and Lorelai is not bringing up something entirely random. 
         \end{itemize}}
        \item[{\scshape rory}]  Yup. 
        \item[{\scshape lorelai}] Right. 
        \item[{\scshape emily}] You didn't read yours? 
        \item[{\scshape lorelai}] Not yet. 
        \item[{\scshape emily}] Ah. 
        \item[{\scshape lorelai}] But you knew that.
        \item[{\scshape emily}] Well.
        \item[{\scshape lorelai}] Hence the smile. 
        \item[{\scshape emily}] Lorelai, you're really being silly. There's no evil plan afoot here. I simply brought up a subject I thought we could all talk about. 
        \item[{\scshape lorelai}] Oh right. 
        \item[{\scshape emily}] [{\sffamily\small Changing subject:}] I'll try another subject---the colour blue is very pleasant, isn't it? 
        \item[{\scshape lorelai}] Mom! Not everybody can wait outside the mailbox for the Chilton newsletter to arrive and then instantly memorize the contents in three seconds.
        {\sffamily\small\begin{itemize}
         \item[\#] Lorelai's goal: explain why she hasn't read the Chilton newsletter yet $\rightarrow$ topic: Lorelai's self-defense 
         \end{itemize}}        
        \item[{\scshape rory}] I'd like to weigh in on the blue colour subject, please. 
        \item[{\scshape emily}] You have your priorities. Far be it from me to question them. 
        \item[{\scshape lorelai}] Just because I don't read the newsletter doesn't mean I don't care about my daughter!
        \item[] {\hfill(I.xi)}
    \end{itemize}
\end{quote}

\paragraph{Example 2.} Now, we revisit our first real example, this time at sentence level (with line break added at the end of each sentence) with attention paid to floor and topic change.

\begin{quote}
    \begin{itemize}
\item[{\scshape emily}] You're being stubborn, as usual.
{\sffamily\small\begin{itemize}
 \item[\#]{Emily's goal: convince Lorelai to do something $\rightarrow$ topic: Lorelai's stubbornness.}
 \end{itemize}}
\item[{\scshape lorelai}] No, Mom, I'm not being stubborn. \\ 
I'm being me! \\ 
The same person who always needed to work out her own problems and take care of herself. \\ 
Because that's the way I was born. \\
That's how I am!
\item[{\scshape emily}] Florence, I'm dripping.  {\sffamily\small\begin{itemize}
 \item[\#]{Emily's goal: have Florence deal with the dripping $\rightarrow$ topic: perm}
 \item[\#]{Emily engages a new character but she does not grab any attention from Lorelai, nor does Florence join the conversation, so no floor change.}
 \end{itemize}}
\item[{\scshape lorelai}] I appreciate what you have done for Rory in paying for this school.
{\sffamily\small\begin{itemize}
 \item[\#]{continuing the previous topic}
 \end{itemize}}
That will not be forgotten.\\ 
You won't let it. \\
But she is my daughter.\\
And I decide how we live, not you.\\
{\sffamily\small [Changing subject:]} Now then, do they validate parking here?
{\sffamily\small\begin{itemize}
 \item[\#]{Lorelai's goal: find out about parking validation on the premise $\rightarrow$ parking}
 \end{itemize}}
    \item[] {\hfill(I.ii)}
    \end{itemize}
\end{quote}

\paragraph{Example 3.} Pay attention to how threads can intervene, and topic is a pragmatic phenomenon.

\begin{quote}
    \begin{itemize} 
\item[{\scshape kirk}] This doesn't smell right. 
{\sffamily\small\begin{itemize}
 \item[\#]{action: fix the egg situation $\rightarrow$ theme: his dining experience}
 \end{itemize}}
\item[{\scshape lane}] Smells fine, Kirk. 
\item[{\scshape kirk}] I think the eggs were bad. 
\item[{\scshape lane}] The eggs are fine, Kirk. 
\item[{\scshape kirk}] Were they cooked in the fish pan? They smell like they were cooked in the fish pan. 
\item[{\scshape lane}] No, the eggs were not cooked in the fish pan. They were cooked in the egg pan. 
\item[{\scshape kirk}] Was the fish pan sitting next to the egg pan? Because perhaps---
\item[] {[\textit{Lorelai walks in the door}.]}
\item[{\scshape lorelai}] {\sffamily\small [Now, everyone look at me:]} I need something with cheese! 
{\sffamily\small\begin{itemize}
 \item[\#]{reading ahead will suggest Lorelai's goal is to find Luke. ``I need something with cheese'' is just like ``Hey guys'', an expression to grab everyone's attention.}
 \end{itemize}}
\item[{\scshape kirk}] Lorelai, smell my eggs. 
{\sffamily\small\begin{itemize}
 \item[\#]{continuing his egg topic}
 \end{itemize}}
\item[{\scshape lorelai}] {Not today, Kirk. {\sffamily\small\begin{itemize}
 \item[\#]{replies to Kirk after she started her own thread}
 \end{itemize}} 
 Hey, where's Luke? I want him to make that breakfast quesadilla thing he made yesterday. }
\item[{\scshape lane}]  Luke's not here. 
\item[{\scshape lorelai}] Where is he? He knows the exact right jack-to-cheddar ratio. 
\item[{\scshape kirk}] He's out there. [\textit{Kirk points out the window.}] 
\item[{\scshape lorelai}]  Where? 
\item[{\scshape kirk}] Over there with Nicole.
    \item[] {\hfill(IV.xvi)}
    \end{itemize}
\end{quote}

\paragraph{Example 4.} This one shows how thread membership can be hard to determine at times. 

\begin{quote}
    \begin{itemize} 
    \item[]{[\textit{Jess enters.}]}
    \item[{\scshape jess}] Morning.
    \item[{\scshape luke}] You're up early.
    \item[{\scshape jess}] {Gotta catch me that worm. \\See ya.}
    \item[{\scshape luke}] Where you off to?
    \item[{\scshape jess}] School.
    \item[{\scshape luke}] This early?
    \item[{\scshape jess}] {I got a lab project going on.\\Me and my team are meeting early.}
    \item[{\scshape luke}] Well, have a good day.
    \item[{\scshape jess}]{If I have a choice.\\{[\textit{to Rory and Lorelai}]} Hey.}
    {\sffamily\small\begin{itemize}
     \item[\#]{Floor change: Jess joins the conversation and attracts the attention from Lorelai and Rory.}
     \end{itemize}}
    \item[{\scshape lorelai}] Good morning.
{\sffamily\small\begin{itemize}
 \item[\#]{Lorelai replies to Jess's ``Hey.''}
 \end{itemize}}
    \item[{\scshape jess}] Talk to you later. [\textit{He and Rory kiss}]
{\sffamily\small\begin{itemize}
 \item[\#]{Jess still has the floor, and there's no obvious change in topic. We have a series of non-topical statements here, and this does not reasonably start a new thread. He, however, does not reply to Lorelai's ``Good morning'' either. So, the most logical reply-to here is his own ``Hey.''}
 \end{itemize}}
    \item[{\scshape rory}] Later.
    \item[]{[\textit{\textit{Jess exits. Lorelai's cell phone rings.}}]}
    \item[{\scshape lorelai}] [\textit{to Rory}] By the way, your boyfriend snores.
{\sffamily\small\begin{itemize}
 \item[\#]{Tricky! See below.}
 \end{itemize}}
    \item[{\scshape rory}] Didn't need to know that. 
    \item[] {\hfill(III.xvii)}
    \end{itemize}
\end{quote}

\noindent It's tricky to determine whether ``By the way, your boyfriend snores.'' is the start of a new thread. Jess's snoring is a new subject. Jess left the room, and no one can pay attention to him anymore, so there's a redistribution of attention. It is also true, however, that most of the thread is composed of semantically non-topical utterances. Since we're reaching the end of scene, this thread would have only two utterances, making it very short. In general, \textbf{we want floor/topic change (the way we define it) to take precedence}, despite some caveats and potential disagreements. Given our definition, we have both topic and floor changes here, so we would like to make ``By the way, your boyfriend snores.'' be the start of a new thread. 

This last example is also an important reminder that ultimately this project is about finding the reply-to relationship (and from there, threads of conversations). It's not about who replies to whom or who is listening. Addressees or participant roles should not be your primary judgement in deciding the reply-to relationship. 

Lastly, many social media and instant messaging apps have this notion of ``thread'' built in: a Twitter or Slack thread usually explores the same topic; if one person explicitly indicates to which previous message they are replying, and then another one replies to that message, those messages naturally form a thread. Threads work in a similar way here.

\subsection{Annotation in action}

\paragraph{Data disclaimer.} Please be aware that our sampled scenes may contain potentially problematic content, such as vulgar, violent, or offensive language in screenplays, or other biases held by individual screenwriters.

\paragraph{Annotation principles.} Here are the general rules for annotation:

\begin{enumerate}
    \item Intuitively, dialogues follow the \textit{basic economic principle}, where $D_n$ replies to $D_{n-1}$. 
    \item A new thread starts when a speaker refocuses other  characters' attention or starts a new topic. 
    \item Use speaker labels, action lines, dialogue turn information to enhance your understanding of the scene.
    \item Always quickly skim through a couple of lines and get a sense of what's happening in the scene before starting to annotate.
\end{enumerate}

\paragraph{Summary of symbols.} For each sentence in a dialogue line, annotate with the following symbols. Next section is on how to use those symbols.

\begin{itemize}
    \item this line is the start of a new thread:
    \begin{itemize}
        \item \verb|T|: \textit{both} floor and topic changes occur, or any signals that indicate the previous conversation was over. Also use this symbol at the start of any scene.
        \item \verb|F|: you're certain only Floor change occurs/can add ``now, everyone listen to me'' at the beginning (or a phrase that serves the same function is actually part of the line)
        \item \verb|P|: you're certain only toPic change occurs/can add ``switching subject'' at the beginnin  (or a phrase that serves the same function is actually part of the line)
    \end{itemize}
    \item \verb|-|: this line replies to the preceding line
    \item \verb|D|$x$: this line replies to sentence $D_x$
    \item symbols for editorial convenience (should be used very sparingly):
    \begin{itemize}
    \item \verb|S|: skip the current sentence, due to significant OCR/parsing errors
    \item \verb|X|: this line requires further discussion for adjudication
    \end{itemize}
\end{itemize}

\begin{figure}
\centering
\begin{subfigure}[t]{0.5\textwidth}
\centering
\includegraphics[width=\textwidth]{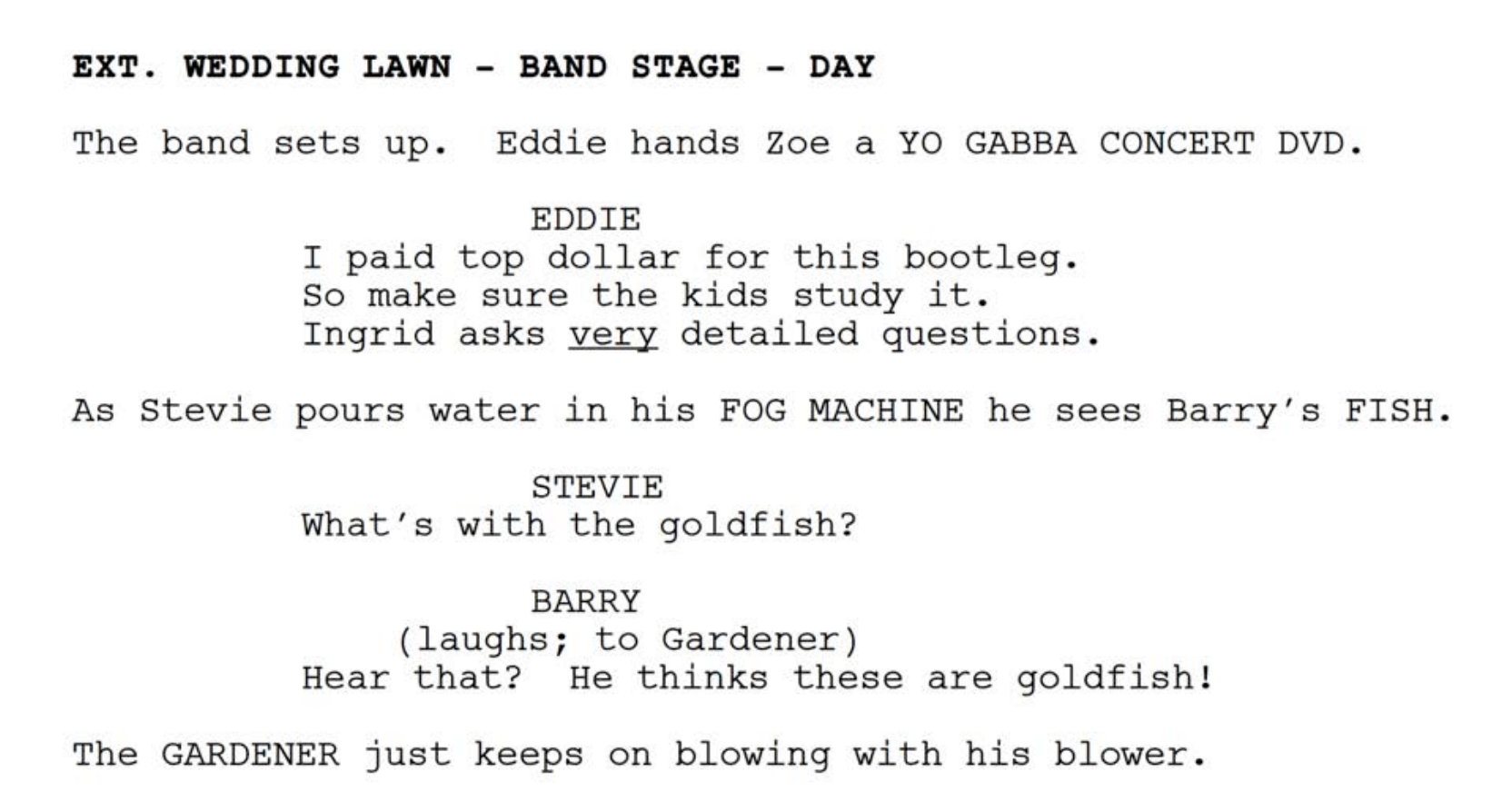} 
\caption{the original pilot script for \textit{The Wedding Band}} \label{fig:timing1}
\end{subfigure}

\begin{subfigure}[t]{0.5\textwidth}
\centering
\includegraphics[width=\textwidth]{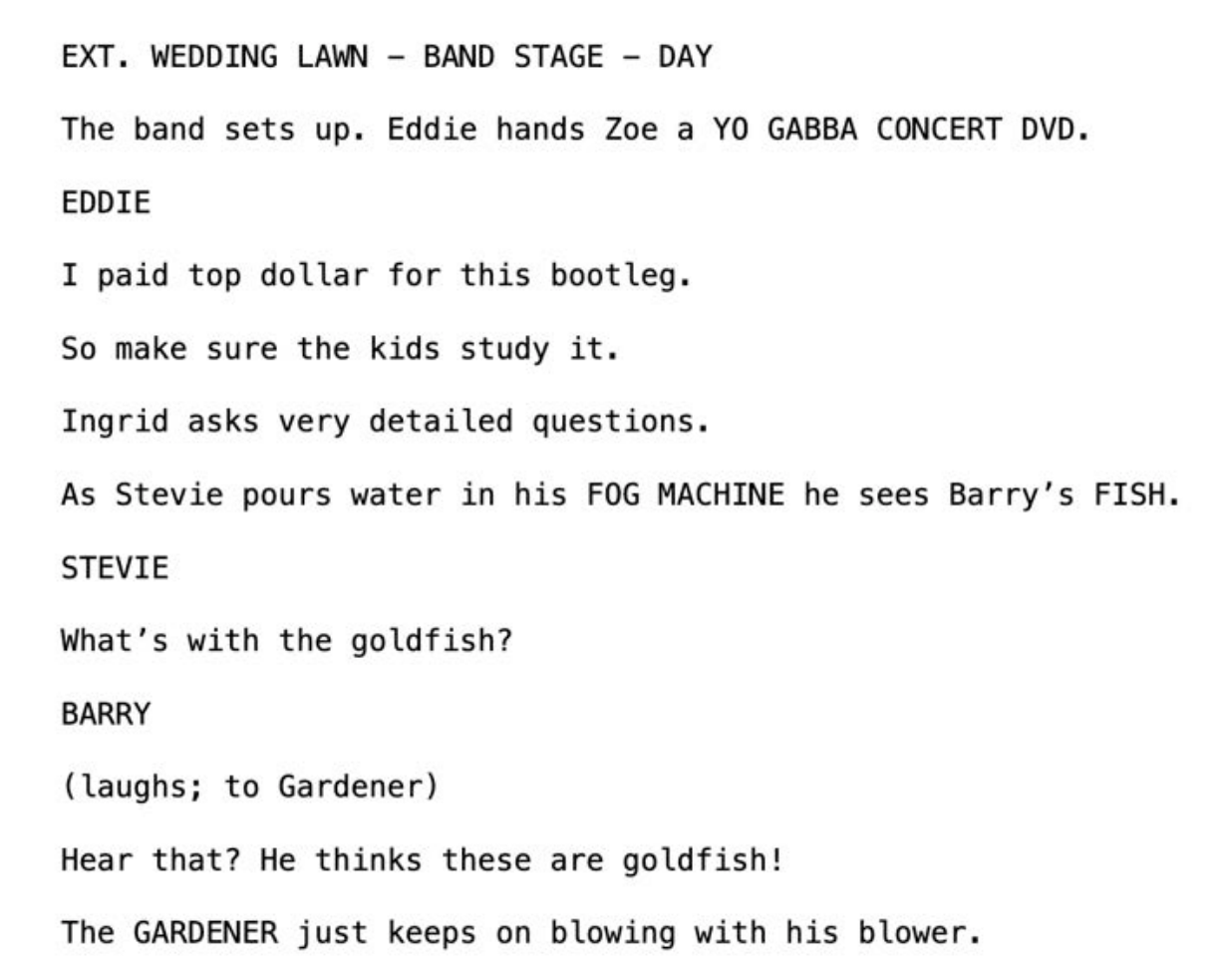} 
\caption{script after OCR} \label{fig:timing2}
\end{subfigure}

\begin{subfigure}[ht!]{0.5\textwidth}
\centering
\includegraphics[width=\textwidth]{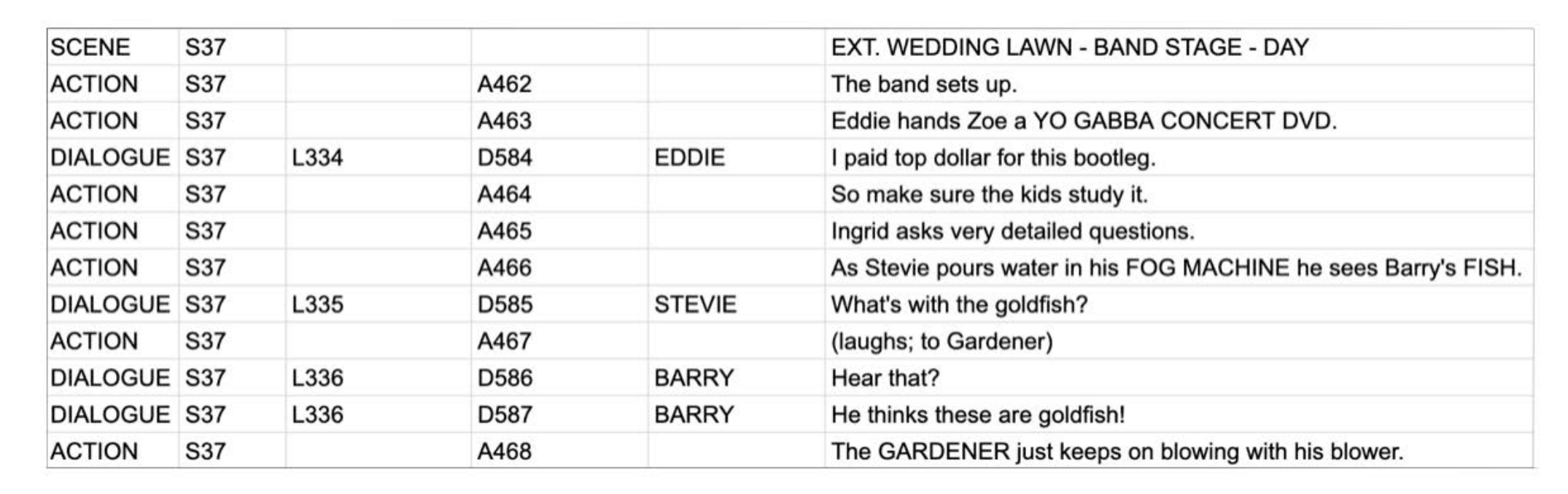} 
\caption{Script after parsing/pre-processing} \label{fig:timing3}
 \end{subfigure}

 \caption{This is an example of typical OCR/parsing errors. It's unclear why Eddie's line is broken into three paragraphs (which is not normal), but since our pipeline relied heavily on line breaks, our parser wrongly recognized Eddie's lines as action statements because you can't really distinguish between a true action statement (``As Steve pours water \dots'') and a dialogue line just by using line breaks. Errors like these are easily fixed, however: supply dummy line turn ID and dialogue IDs, and annotate accordingly.}\label{fig}

\end{figure}

\paragraph{Handling parsing/OCR errors} Fig.~\ref{fig} shows you what kind of parsing/OCR related errors you might encounter and why they are there. If you spot an obvious/easily fixable OCR or parsing error, please correct it. If you suspect you've spotted an error of any kind, you could take a look at the original \verb|txt| or \verb|pdf| file. It also comes with experience. After you're sure there are errors, here's how you fix them:

First, if you are turning an action line into a dialogue line, supply dummy dialogue turn ID and dialogue ID. Our suggestion is something like \verb|La| and \verb|Da|. We do so because those lines you are rescuing might become parent utterances of UOIs to come, in which case you can annotate with \verb|Da|.

Second, you might need to turn a dialouge line into an action line:

\medskip

\noindent{\small \begin{tabular}{rrrrrl}
  \hline
  turn id & line id & speaker & line \\
  \hline
  L371 & D470 & THE BATHROOM & Where \dots \\
  \hline
\end{tabular}\medskip}

\noindent Many entities are singled out printed in uppercase, such as THE BATHROOM here. When they appear alone in the line, there's no way to distinguish them from a regular speaker label (we also don't want to simply exclude locations, because e.g., MAN IN THE STREET is a well-formed speaker label). To correct this, simply change the dialogue ID to \textbf{A} Changing line type is NOT necessary (saves you two seconds, which add up). Since we don't really use action IDs for annotation, it's not necessary to add them. Removing dialogue turn ID or speaker label is optional. This is what that row should be:

\medskip

\noindent{\small \begin{tabular}{rrrrl}
  \hline
  turn id & line id & speaker & line \\
  \hline
  L371 & \textbf{A} & THE BATHROOM & Where \dots \\
  \hline
\end{tabular}\medskip}

Third, if a line is now empty after your correction, or if you spot a line that does not contain an action statement or dialogue line but has information about the script or purely logistical information (like \verb|Untitled Project (04/12/22)|, \verb|CBS Studio Production|), put \verb|S| (skip). Or long-tail errors: screenplays to \textit{Star Trek} movies put all dialogues in Klingon in parentheses, and they will be parsed as action lines. It's impossible to find all of them through regular expressions, and we don't need our model to see them.

Fourth, pay attention to ellipses and em dashes: IMSDB/Scriptbase-J can use `[sentence] . . .' (with leading space), `[sentence]. . .' (without), and there can be space between each dot (`. . .' vs `...'). Em-dashes can be ` - ' (dash separated by space), `--' (two dashes, no space surrounding), or `-' (that looks just like two words being linked together. There's no easy way for us can clean and normalize that in our pipeline, and in some cases they interfere with the semantics. So correct those too.

\subsubsection{Questions to ask while annotating}

\begin{enumerate}
    \item This is the beginning of a scene. 
    \begin{itemize}
       \item Put \verb|T|.
       \item Read a couple of lines ahead and gain a sense of:
       \begin{itemize}
        \item Why does the character speak at all? 
        \item What do they want? 
        \item Who has the floor?
        \end{itemize}
    \end{itemize}
    \item This is a new sentence:
    \begin{enumerate}
        \item Can $D_{n-1}$ be the sensible reply to $D_n$?
        \begin{itemize}
            \item If so, put \verb|-|.
        \end{itemize}
        \item If not, what previous line leads to this line? Is there any previous line that triggers (or, gives the necessary context for us to understand) this UOI?
        \begin{itemize}
            \item If there's one, put the utterance ID.
        \end{itemize}
       \item If not, is there a topic/floor change?
       \begin{enumerate}
           \item Is there any new intent or desire being expressed? Can I insert ``switching subject'' at the beginning of UOI?
            \begin{itemize}
                \item If so, put \verb|P| (toPic).
            \end{itemize}           
           \item Is there a character replacing another one as the center of attention? Can I insert ``Now, look at me/listen to me'' at the beginning of UOI?
            \begin{itemize}
                \item If so, put \verb|F| (Floor).
            \end{itemize}           
           \item Do floor and topic changes happen at the same time? Or are there other signals that make you think the previous conversation is over (someone has left the room, or other observable contextual changes)?
            \begin{itemize}
                \item If so, put \verb|T| (Thread).
            \end{itemize}
       \end{enumerate}
       \item If none applies, this seems to be an edge case. Follow the following steps:
       \begin{enumerate}
           \item Is this a dialogue line wrongly parsed as action line? 
           \begin{enumerate}
               \item If so, change the line ID from \verb|A|$x$ to \verb|D|$x$.
               \item Correct it. Supply dialogue turn ID and speaker label to match the format of a regular dialogue line. Use the original \verb|txt| or \verb|pdf| file if necessary.
               \item If there are any empty lines (for example: you moved an obvious speaker label from one line to the ``speaker'' column of the current line) or lines that contain redundant information (for example: the content of this line and is copied to the previous line, which made the previous line a complete dialogue line) or information irrelevant to the task at hand (for example: ``PILOT DRAFT \#4''), put \verb|S| (Skip)
           \end{enumerate}
           \item Is this an action line wrongly parsed as a dialogue line?
           \begin{enumerate}
               \item If so, change the line ID from \verb|D|$x$ to \verb|A|.
               \item Correct it.
               \item Put \verb|S| (Skip) when you see fit.
           \end{enumerate}
           \item Are there multiple wrongly parsed lines, or any OCR/parsing errors that takes more than \textbf{TWO MINUTES} to fix?
           \begin{itemize}
               \item If so, don't spend time fixing all that. 
               \item Put \verb|S|. 
               \item Take a note of the corpus name and title slug (e.g., \verb|[tvpilot, the-wedding-band]|). We'll see if we want to remove that title from the corpus altogether later.
           \end{itemize}
           \item Is there no logical parent utterance, but you can't say there's any change in distribution of attention or switch in subject? 
           \begin{itemize}
               \item If so, put \verb|X| and move on. We can discuss later.
           \end{itemize}
       \end{enumerate}
    \end{enumerate}       
\end{enumerate}

\subsubsection{Post-processing} 

We will be post-processing all annotations to: 

\begin{itemize}
    \item convert \verb|-|'s into the correct ID's
    \item change \{\verb|T, P, F|\} into \verb|Tn|: We will NOT distinguish between \verb|T|, \verb|P|, or \verb|F| for modeling, and they will be converted to \verb|T1|, \verb|T2|, etc. 
    \item delete any lines tagged \verb|S| (Skip)
    \item correct any non-conventional dialogue/action line to account for individual fixes: We will create a mapping between old and new line ID's: e.g.,~\verb|D45, Dg, D46|, where \verb|Dg| was added during annotation to fix a parse/OCR error, will become \verb|D45, D46, D47|. 
\end{itemize}

\end{document}